\documentclass[10pt,twocolumn,letterpaper]{article}

\usepackage{wacv}
\usepackage{times}
\usepackage{epsfig}
\usepackage{graphicx}
\usepackage{amsmath}
\usepackage{amssymb}
\usepackage[accsupp]{axessibility}  
\newcommand{\bfu}[1]{\textcolor{black}{#1}}

\def\wacvPaperID{0348} 

\wacvfinalcopy 

\ifwacvfinal
\fi

\ifwacvfinal
\usepackage[breaklinks=true,bookmarks=false]{hyperref}
\else
\usepackage[pagebackref=true,breaklinks=true,colorlinks,bookmarks=false]{hyperref}
\fi

\begin{document}

\title{A Deep Insight into Measuring Face Image Utility with General and Face-specific Image Quality Metrics}

\author{Biying Fu$^{1}$, Cong Chen$^{1}$, Olaf Henniger$^{1}$, Naser Damer$^{1,2}$
\\
$^{1}$Fraunhofer Institute for Computer Graphics Research IGD,
Darmstadt, Germany\\
$^{2}$Department of Computer Science, TU Darmstadt,
Darmstadt, Germany\\
Email: {biying.fu@igd.fraunhofer.de}
}


\maketitle

\ifwacvfinal
\thispagestyle{empty}
\fi

\begin{abstract}

Quality scores provide a measure to evaluate the utility of biometric samples for biometric recognition. Biometric recognition systems require high-quality samples to achieve optimal performance. This paper focuses on face images and the measurement of face image utility with general and face-specific image quality metrics. While face-specific metrics rely on features of aligned face images, general image quality metrics can be used on the global image and relate to human perceptions. In this paper, we analyze the gap between the general image quality metrics and the face image quality metrics. Our contribution lies in a thorough examination of how different the image quality assessment algorithms relate to the utility for the face recognition task. The results of image quality assessment algorithms are further compared with those of dedicated face image quality assessment algorithms. In total, 25 different quality metrics are evaluated on three face image databases, BioSecure, LFW, and VGG\-Face2 using three open-source face recognition solutions, SphereFace, Arc\-Face, and FaceNet. Our results reveal a clear correlation between learned image metrics to face image utility even without being specifically trained as a face utility measure. Individual handcrafted features lack general stability and perform significantly worse than general face-specific quality metrics. We additionally provide a visual insight into the image areas contributing to the quality score of a selected set of quality assessment methods.  
\end{abstract}

\section{Introduction}

Face recognition (FR) has gained high user acceptance due to the convenience and high accuracy \cite{DBLP:journals/corr/abs-2109-09416}. Commercial products, such as Google Pay \cite{janssen2018google} and Apple Pay \cite{liu2019apple} integrate FR for user authentication to make digital payment easier and more secure. The automatic FR on border control further reduces the workload of border officers and accelerates the process. The widespread use of FR solutions are only possible due to the recent advances made in deep-learning based FR algorithms.

The quality of a face image directly affects the performance of the underlying FR systems. Face image quality assessment (FIQA) assigns a quality score to an input face image expressing its utility to support a correct outcome of an FR decision. The term ``utility'' considered in this work is based on the definition in ISO/IEC 29794-1 \cite{ISOIEC29794-1} relating the quality score (QS) to biometric performance. The output of a specific FIQA algorithm may depend on a specific FR system used for training. This entails that the face image utility is conditioned on both, the face image and the specific FR system. Such metrics should preferably be indifferent to the FR solution and work in general.

In contrast to FIQA, image quality assessment (IQA) works more generalized but depends on the subjective viewpoints of the end-users. IQA algorithms link the quality score (QS) of an image with the mean opinion scores (MOS) assigned by multiple human observers. Popular subjective IQA databases used for training IQA algorithms are e.g.,~LIVE \cite{sheikh2005live}, TID2008 \cite{ponomarenko2009tid2008}, and TID2013 \cite{ponomarenko2015image}, each containing different types of visible image distortions, such as additive noise, compression errors, \bfu{blurring,} and contrast changes. Trained IQA methods as in \cite{bosse2017deep, kang2014convolutional, ma2017end} predict image quality without a reference image.

FIQA and IQA algorithms have been developed as two separate streams. Only limited research work can be found studying the relationship or the possible interoperable use between these two research streams as in \cite{dutta2014bayesian,best2018learning,dutta2015predicting}. Natural image statistics like BRISQUE, NIQE, and PIQE are sometimes used as baselines to emphasize the strong performance of learned face-specific quality metrics such as in \cite{meng2021magface, terhorst2020ser}. In this paper, we investigate IQA and FIQA algorithms to provide a deep insight into their role in face recognition systems. In addition, based on the utility conditioned on the FR performance for specifically designed experiments, we aim to answer questions related to the correlation of both, the general image quality metrics and the face-specific quality metrics, to the face image utility.

To address these research challenges, we build our take-home conclusions based on the evaluation of 25 different quality metrics, which can be categorized into four groups: 1) general image quality measures, 2) handcrafted image quality measures, 3) face related handcrafted measures, and 4) learned face utility measures. The paper is structured as follows: \bfu{In Section~\ref{sec2}, we show the high-level progress made in the fields of IQA and FIQA methods. In Section~\ref{methods_fiqa} and Section~\ref{methods_iqa}, we introduce selected algorithms from both research domains to draw comparisons by conducting experiments (in Section~\ref{evaluation}) specifically designed to address three research questions, followed by the results, analysis and a visual interpretation (Section~\ref{sec:visualization}) to emphasize our findings. Section~\ref{conclusion} recapitulates the quintessence of the paper.} 

\section{Related work}\label{sec2}

Perceptual IQA assesses the distortion and degradation on the visual material, such as compression, white noise (WN), or Gaussian blur (GB). These types of distortion can also cause the face image utility to drop. The research domain of IQA is sub-divided into full-reference, reduced-reference, and no-reference IQA. Full-reference IQA algorithms require an original image for comparison, e.g., peak signal-to-noise ratio (PSNR) \cite{huynh2008scope}, and structural similarity index measure (SSIM) \cite{ssim}. In contrast to these methods, the Reduced-reference IQA algorithms provide a solution for image quality estimation where only partial information is accessible e.g., in \cite{rehman2012reduced,DBLP:conf/icpr/DamerSN14, wu2013reduced, li2009reduced}. 

Usually one does not have a reference image directly to verify the comparison score in an FR system. Therefore no-reference IQA is the most similar case to FIQA. Dutta et al. proposed in \cite{dutta2014bayesian, dutta2015predicting} to utilize a Bayesian framework to link feature-based FIQA (e.g.,~focus, pose, and illumination direction) to predict FR performance, but did not consider IQA methods in general. However, to our knowledge, only limited previous works explore the relation between no-reference IQA methods and the face utility. 

ISO/IEC TR 29794-5 \cite{ISOIEC29794-5} defines image-level features, such as illumination symmetry, inter-eye distance, blur, sharpness, and person-dependent features such as wearing beards, eyeglasses, or eyes/mouth closed/opened as required in \cite{ICAO18}. These features are assumed to have a direct impact on the face utility. While these generated features are explainable and directly connected with image qualities, several current FIQA algorithms \cite{hernandez2019faceqnet,terhorst2020ser,meng2021magface} rely on methods using end-to-end learning. This trend is due to the significant improvement of deep-learning-based (DL-based) FR solutions in both industry and academic fields. Only very few works focus on relating FIQ with morphed faces \cite{9548302} or face parts \cite{9548297} to enhance interpretability of these measures.  

\section{Face image quality assessment algorithms}\label{methods_fiqa}

\bfu{We selected six DL-based FIQA methods as they demonstrated state-of-the-art performances and are based on various training and conceptualization strategies. They can be grouped into categories of either supervised on quality labels, e.g., FaceQnet \cite{hernandez2019faceqnet} or unsupervised methods, e.g., rankIQ \cite{chen2015face}, MagFace \cite{meng2021magface}, SER-FIQ \cite{terhorst2020ser}, PFE \cite{DBLP:conf/iccv/ShiJ19}, and SDD-FIQA \cite{DBLP:journals/corr/abs-2103-05977}.} We further examine explainable image-level features as proposed by ISO/IEC TR 29794-5 \cite{ISOIEC29794-5}.

\subsection{Deep learning-based FIQA Methods}

\textbf{RankIQ} \cite{chen2015face} is a model trained to assess face utility using a ranking-based approach. \bfu{The author is inspired by the premise that the quality of facial images cannot be quantified absolutely and is easier to be considered in a relative manner. This method was trained using three databases with varying qualities. The training is a two-stage process. While stage I learns to map the individual face image features (e.g., HoG, Gabor, LBP, and CNN features) to first level rank weights, the stage II maps the learned feature scores to a final normalized quality score by using kernels.}

\textbf{FaceQnet} \cite{hernandez2019faceqnet} \bfu{by Hernandez-Ortega et al. is a supervised, and DL-based method trained on VGGFace2 \cite{Cao18} database. We used FaceQnet v2 which is the most recent version of FaceQnet} \cite{hernandez2020biometric}\footnote[2]{FaceQnet Github: https://github.com/uam-biometrics/FaceQnet}. The BioLab-ICAO framework \cite{Fer12} was used to select ICAO-compliant high-quality reference images. FaceQnet then fine-tuned a pre-trained FR base-network (RseNet-50 \cite{he2016deep}) and the successive regression layer on top of the feature extraction layers to associate an input image to a utility score. The target utility score is the normalized similarity score between the face image and a high-quality mated reference image. 

\textbf{SER-FIQ} \cite{terhorst2020ser} is an unsupervised DL-based FIQA method based on a stochastic method applied on face \bfu{representations.} This method mitigates the need for any automated or human labeling. The face image was passed to several sub-networks of a modified FR network by using dropout. Images with high-utility are expected to possess similar face representations resulting in low variance. Therefore, the proposed method linked the robustness of face embeddings directly with face utility. \bfu{In our study, we used the method finetuned with ArcFace loss \cite{deng2019arcface} using ResNet-100 \cite{he2016deep} base architecture trained on MS1M database \cite{guo2016ms} named the SER-FIQ (on ArcFace) method.}

\bfu{Shi et al. proposed the \textbf{Probabilistic Face Embeddings (PFEs)}~\cite{DBLP:conf/iccv/ShiJ19}, which represents each face image as a Gaussian distribution in the latent space. The model is trained to maximize the mutual likelihood score of all genuine pairs equally to map distorted input face to its genuine latent space. The mean of the Gaussian represents the most likely features and the variance indicates the uncertainty in the features values. In this paper, we refer the variance to face utility, where a low quality face image possess a larger uncertainty in the latent embedding space.}

\bfu{\textbf{MagFace} \cite{meng2021magface} by Meng et al.~is a recently developed method to derive both the face representation and the face image quality from calculating the magnitude of the face embedding. They extended the ArcFace loss \cite{deng2019arcface} by an adaptive margin regularization term to further enforce the easily recognizable samples towards the class center and the hard samples further away. The face utility is inherently learned through this loss function during training. The magnitude of the feature vector is proportional to the cosine distance to its class center and is directly related to the face utility.}

\bfu{\textbf{SDD-FIQA} by Ou et al.~\cite{DBLP:journals/corr/abs-2103-05977} is a novel unsupervised FIQA method that incorporates Similarity Distribution Distance for FIQA. The method leverages the Wasserstein Distance (WD) between the inter-class samples (Neg-Similarity) and the intra-class samples (Pos-Similarity). The FR model uses ResNet-50 \cite{he2016deep} trained on MS1M database to calculate the positive samples and negative samples distributions. This WD metric is used as the quality pseudo-labels to train a regression network for quality prediction with Huber loss.}

\subsection{FIQA metrics based on Handcrafted Features }

We chose eight representative features to extract information from a face image according to the ISO/IEC Technical Report \cite{ISOIEC29794-5}. We aim to relate these individual features to assess the utility of the underlying FR systems. Features derived from the spatial domain are blur, contrast, mean, luminance, sharpness, lighting symmetry, and exposure. They depend on the pixel intensity and its statistical distribution. Inter-eye distance measures the distance between two eye middle points from the original image and is assumed to be directly related to face image utility. The set of features implemented are seen in the legend of Figure~\ref{fig:erc_001_hc}.

\section{Image quality assessment algorithms}\label{methods_iqa}

\bfu{In this section, we present ten IQA methods, which can be categorized into: (1) model-based, (2) CNN-based, (3) multi-task learning-based, and (4) rank-based approaches. In our experiment, we associate these IQA metrics to face image utility. These methods are not specifically designed to assess face image utility for face recognition.}

\subsection{Model-based IQA methods}

\bfu{This IQA category tries to build a model to assess the general image quality based on natural image statistics. Here, we cite three methods 
\textbf{BRISQUE} \cite{mittal2012no}, \textbf{NIQE} \cite{mittal2012making}, and \textbf{PIQE} \cite{venkatanath2015blind} that all based on studying the deviation from the general statistics of natural images. This statistical model is derived from the finding by Rudermann \cite{ruderman1994statistics} that natural scene images have a luminance distribution similar to a normal Gaussian distribution. The degree of deviation from the normal Gaussian distribution relates to the degradation in image quality. While NIQE and PIQE are both opinion-unaware methods without the need for human-rated scores, BRISQUE requires training with human opinion scores. We selected these three methods, as they were used as baseline for performance comparison in FIQA methods like SER-FIQ and MagFace. The results demonstrated on the error vs. reject characteristic curve showed that these three methods are inferior compared to the SER-FIQ algorithm concerning face quality assessment.}

\subsection{CNN-based IQA algorithms}

\bfu{CNNIQA\cite{kang2014convolutional} and DeepIQA \cite{bosse2017deep} are both methods that leverage CNNs in the base layers to automatically extract spatial image features from the input image avoiding generating handcrafted features as opposed to methods in the previous sub-section. While \textbf{CNNIQA} \cite{kang2014convolutional} used a shallow net with only one convolutional layer, the \textbf{DeepIQA} \cite{bosse2017deep} proposed by Bosse et al. applied a deeper structure with multiple stacked convolutional layers. Compared to CNNIQA, the increased model capacity allows the model to deal with more complex and colored images. Both methods are derived from patch-wise quality score where local metrics are aggregated to a global metric resulting in a whole image quality estimate. Although other methods can be based on CNNs, we refer in this category to methods with single-task learning (i.e. CNNs solely trained to estimate a quality value).}

\subsection{Multi-task learning-based IQA algorithms}

\bfu{Training multiple tasks within one network structure often enhance the network ability in performing one specific task \cite{DBLP:conf/icml/Caruana96}. MEON \cite{ma2017end} and DBCNN \cite{DBLP:journals/tcsv/ZhangMYDW20} are both mutlitask-learning-based approaches trained to classify the type of image distortions on the one hand to enhance the ability of estimating the image quality on the other hand in one combined network. \textbf{MEON} \cite{ma2017end} used a shared network as base structure with two parallel networks added on top, one for identifying the type of distortion (Subtask I) and one for predicting the quality of a given input image (Subtask II). Subtask I can benefit from generating large quantity of synthetic data at low cost. Similar concept is used for \textbf{DBCNN}~\cite{DBLP:journals/tcsv/ZhangMYDW20} where two CNNs were trained at the base, one on large-scale synthetically generated databases, while the other focuses on the classification network pre-trained on ImageNet \cite{DBLP:conf/cvpr/DengDSLL009} for extracting more authentic distortions. Finally, the features from both CNNs are pooled bilinearly into a unified representation for a final quality prediction.}

\subsection{IQA based on ranked image pairs}

\bfu{Since image quality is not homogeneously quantifiable, the mean opinion score (MOS) from human operators is considered a useful metric. However, the process of obtaining MOS is tedious. Hence, using relative comparison without quantifiable metric simplifies the design of IQA metrics. UNIQUE~\cite{DBLP:journals/tip/ZhangMZY21}, rankIQA \cite{liu2017rankiqa}, and dipIQ \cite{7934456} are methods that use ranked image pairs to avoid the absolute scale of image quality. Large databases with ranked image pairs can be generated at low cost synthetically as in \cite{7934456,liu2017rankiqa} or assembled as in \cite{DBLP:journals/tip/ZhangMZY21}. \textbf{dipIQ} \cite{7934456} leveraged a two-stream networks trained on quality-discriminative pairs (DIP) of images. Each DIP is associate with a degree of perceptual uncertainty. \textbf{RankIQA} \cite{liu2017rankiqa} used a Siamese network to train on pairs of ranked inputs. Subsequently, the trained Siamese Network can be used to teach a traditional CNN that estimates the image quality using only one single input. \textbf{UNIQUE} ~\cite{DBLP:journals/tip/ZhangMZY21} used a deep neural network structure. The training is based on pairs extracted from multiple IQA database to acquire the MOS score and the corresponding variance. The network training uses the fidelity loss to optimize the DNN over a large number of such image pairs and the hinge loss to regularize the uncertainty estimation during optimization. The network learns the mean and the variance of an input image which represent the quality score and the uncertainty respectively.} 

\bfu{An overview  of  the  used  methods are found in Table \ref{tab:methods}, where the previous works in both categories of IQA and DL-based FIQA are introduced in terms of its year of publication, categories, sub-categories, and way of training.}

\begin{table}
\resizebox{\columnwidth}{!}{%
\begin{tabular}{l|l|l|l|l}
\hline
Method   & Conference, Year & Category & Sub-category        & Un-/supervised  \\ \hline
RankIQ \cite{chen2015face}  & IEEE SPL, 2015   & FIQA     & Feature Fusion  & unsupervised    \\ \hline
FaceQnet \cite{hernandez2020biometric} & arXiv preprint, 2020  & FIQA     & ResNet-50       & supervised      \\ \hline
SER-FIQ \cite{terhorst2020face}  & CVPR, 2020       & FIQA     & ResNet-100      & unsupervised    \\ \hline
PFE \cite{DBLP:conf/iccv/ShiJ19} & ICCV, 2019       & FIQA     & 64-layer CNN    & unsupervised    \\ \hline
MagFace \cite{meng2021magface}  & CVPR, 2021       & FIQA     & ResNet-100      & unsupervised    \\ \hline
SDD-FIQA \cite{DBLP:journals/corr/abs-2103-05977} & CVPR, 2021       & FIQA     & ResNet-50       & unsupervised    \\ \hline
BRISQUE \cite{mittal2012no}  & IEEE IP, 2012    & IQA      & Model-based     & supervised      \\ \hline
PIQE \cite{venkatanath2015blind}    & NCC, 2015        & IQA      & Model-based     & unsupervised      \\ \hline
NIQE \cite{mittal2012making}    & IEEE SPL, 2012   & IQA      & Model-based     & unsupervised      \\ \hline
CNNIQA \cite{kang2014convolutional}  & CVPR, 2014       & IQA      & CNN-based       & supervised      \\ \hline
DeepIQA \cite{bosse2017deep} & IEEE TIP, 2018   & IQA      & CNN-based       & supervised      \\ \hline
MEON \cite{ma2017end}    & IEEE TIP, 2018   & IQA      & Multitask-based & unsupervised \\ \hline
DBCNN \cite{DBLP:journals/tcsv/ZhangMYDW20}   & AICCSA, 2020     & IQA      & Multitask-based & semi-supervised \\ \hline
dipIQ \cite{7934456}   & IEEE TIP, 2017   & IQA      & Ranking-based   & unsupervised    \\ \hline
RankIQA \cite{liu2017rankiqa} & ICCV, 2017       & IQA      & Ranking-based   & unsupervised    \\ \hline
UNIQUE \cite{DBLP:journals/tip/ZhangMZY21}   & IEEE TIP, 2021   & IQA      & Ranking-based   & supervised    \\ \hline
\end{tabular}
}
\caption{Table provides an overview of the used methods in both categories of IQA and DL-based FIQA.}
\label{tab:methods}
\end{table}

\section{Experimental Evaluation}
\label{evaluation}

In this section, we first introduce the used databases, FR solutions, and evaluation metrics. Then, the experimental results will be presented in a comprehensible manner. 

\subsection{Face Image Databases}

Three face image databases are used to link the scores estimated by FIQA methods in Section~\ref{methods_fiqa} and IQA methods in Section~\ref{methods_iqa} with the face image utility. 

The \textbf{BioSecure} \cite{ortega2009multiscenario} DB contains face images of 210 subjects with only frontal views and highly controlled quality data. We chose this database for the following reasons: 1) it represented the controlled and collaborative use case scenario relevant for border checks and identity documents (ISO/IEC 19794-5), and 2) it was reported for face quality in (\cite{hernandez2019faceqnet} ICB2019). We conducted a series of experiments on this database. However, the results show that due to the high-quality images, the FR systems perform almost perfectly.

The Labeled Faces in the Wild (\textbf{LFW}) DB \cite{LFWTech} is a widely used standard benchmark for automatic face verification. It contains in total 13233 images from mostly uncontrolled scenarios. We chose this database as it is used in the FIQA methods \cite{terhorst2020ser, meng2021magface}. However, this database is strongly imbalanced regarding the number of images for each subject. 

Similarly, \textbf{VGGFace2} \cite{Cao18} test contains 500 subjects. We chose VGGFace2 due to its large variety in quality distribution which can be considered a challenging database. The images have diverse and complex acquisition conditions. To manage the heavy computation due to the large database, we selected a representative subset by randomly choosing 30 out of 300 images from each subject to perform the 1:1 verification task. 

The MTCNN framework \cite{zhang2016joint} is used to detect, align, and crop the input face images to a fixed size of $260\times260$ pixels. The set of used images are subsequently adapted to the input size of each of the used networks.

\subsection{Face recognition solutions}

Three open-source academic FR solutions are used to derive the face embeddings to perform the verification task. 

We chose \textbf{Facenet}~\cite{schroff2015facenet} because it is one of the first FR solutions based on deep CNN structures using inception resnet as backbone. Triplet loss and center loss are used to facilitate the training. The accuracy reported on the LFW DB is 99.63\%$\pm$0.09 and on YouTube Faces DB (YTF)~\cite{wolf2011face} is 95.12\%\%$\pm$0.39.

\textbf{SphereFace} \cite{liu2017sphereface} used a 64-CNN layers trained on CASIA-WebFace \cite{DBLP:journals/corr/YiLLL14a}. We chose this FR model as it achieved a competitive state-of-the-art verification accuracy on LFW DB to 99.42\% and YTF DB to 95.0\%.

\textbf{ArcFace} \cite{deng2019arcface} is trained on ResNet-100 \cite{han2017deep} using the MS1M dataset \cite{guo2016ms}. The loss function further uses additive angular margin to improve the discriminative power of the FR model. This model is chosen due to its improved accuracy on LFW 99.83\% and YTF DB 99.02\%.

\subsection{Evaluation metric}

The evaluation metric used is the error vs.~reject characteristic (ERC) \cite{GT07}. 
\bfu{The ERC shows the relative performance when rejecting different ratios of the evaluation data with the lowest error. With a "better" face utility estimation, the face verification error should strongly decrease when rejecting more low-quality data.  
In our presented ERC, we show the false non-match rate (FNMR) at different ratios of rejected (low quality) images.
The presented FNMR is the FMR1000, i.e. the FNMR at false match rate (FMR) value of 0.1\% as recommended for border operations by Frontex \cite{frontex2015best}.
For a well-functioning face utility estimation, the FNMR is expected to go down as the ratio of discarded (rejected) images increases. }

For LFW DB, we used the test protocol as reported in \cite{huang2014labeled} to balance the database. The original database contains 5749 subjects, but only 1680 subjects have two or more images. BioSecure and VGGFace2 include already a balanced database and are used in verification scenarios comparing every image with every other image. 

\bfu{To quantitatively represent the correlation between quality estimation methods and categories, we calculate the samples overlap ration between the samples of the lowest quality (10\% of the data) between every pair of quality estimation methods, and the same for the 10\% of the highest quality. A large overlapping ratio indicates a larger reasoning similarity between the considered pair of methods.}


Table~\ref{tab:top_perform} provides an overview of the three top-performing methods compared in sub-groups of DL-based FIQA, feature-based FIQA, and learned IQA methods, for all two face image databases (LFW, and VGGFace2) at FMR1000. The best performing three methods in each setup are emphasized in bold. We report the two reject ratios at 20\,\% and 40\,\%. The full table including the results of all methods is available in the supplementary material for a more comprehensive study.

\subsection{Results on face image databases}

This section presents the evaluation results based on three database and three SOTA FR solutions. The results and discussions are divided into answering separate research questions regarding the effect of IQA and FIQA metrics on the face image utility.

The BioSecure DB is a high-quality database with a controlled capture environment. Due to the consistent high-quality of the images in the DB, the ERC curve shows a very low FNMR and does not reveal any apparent changes across different reject ratios. Since all FR systems perform almost perfectly on the entire database, no interesting observations can be found. However, it was essential to include this database as it represents the controlled (passport-like) data presentation scenario.
Therefore, we only included the results in the supplementary materials for completeness.

\paragraph{Q1: How does FIQA metrics using handcrafted features correlate to face image utility?}

\begin{figure*}
\centering
\includegraphics[width=.95\linewidth]{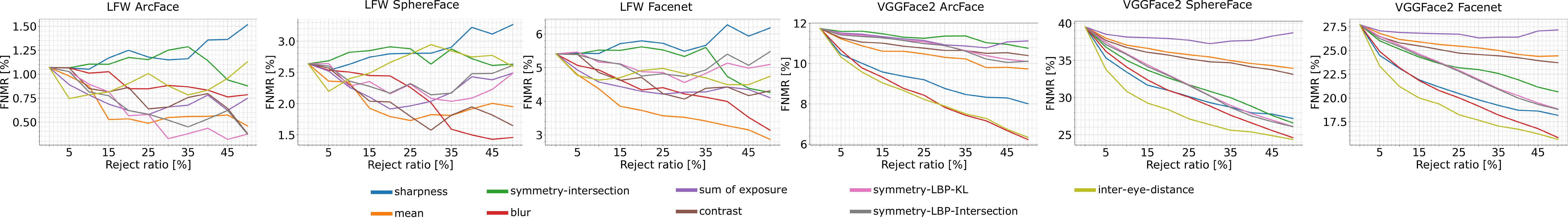}
\caption{Error vs.~reject characteristics at FMR1000 for handcrafted image quality measures and a face-related handcrafted measure like inter-eye distance. The rows reveal the ERC results for different face embeddings on LFW and VGGFace2. Inter-eye distance performed well on VGGFace2 using original images, while sharpness and blur are well performed for aligned images. However, individual feature contributes inconsistently across different settings, while a fusion could be more promising.} 
\label{fig:erc_001_hc}
\end{figure*}

Figure~\ref{fig:erc_001_hc} shows the ERCs at FMR1000 using handcrafted image features as quality metrics. Observing the results on VGGFace2, the inter-eye distance feature seems to be most effective in selecting high-utility face images. However, inter-eye distance cannot obtain consistent results on all DBs. For LFW at FMR1000 with ArcFace embeddings at the reject ratio of 20\%, the mean-feature with an FNMR=0.532\% almost halved the error rate compared to the inter-eye distance feature with an FNMR=0.903\%. This may be due to the fact that handcrafted features such as the inter-eye distance is naive and assume the images are not further processed after capture (not scaled, scanned, etc.) and that the capture device is consistent (the same amount of pixels corresponds to same quality). These strict assumptions do not apply to a database collected from various sources such as LFW. The data collection process of LFW may undergo special processing. Hence it will affect the performance of handcrafted features. In LFW, no single handcrafted feature excels the others and the changes in the error performance are less obvious. The total reduction in error for handcrafted image features accounts less compared to FIQA methods.

\begin{figure*}
    \centering
    \includegraphics[width=.9\linewidth]{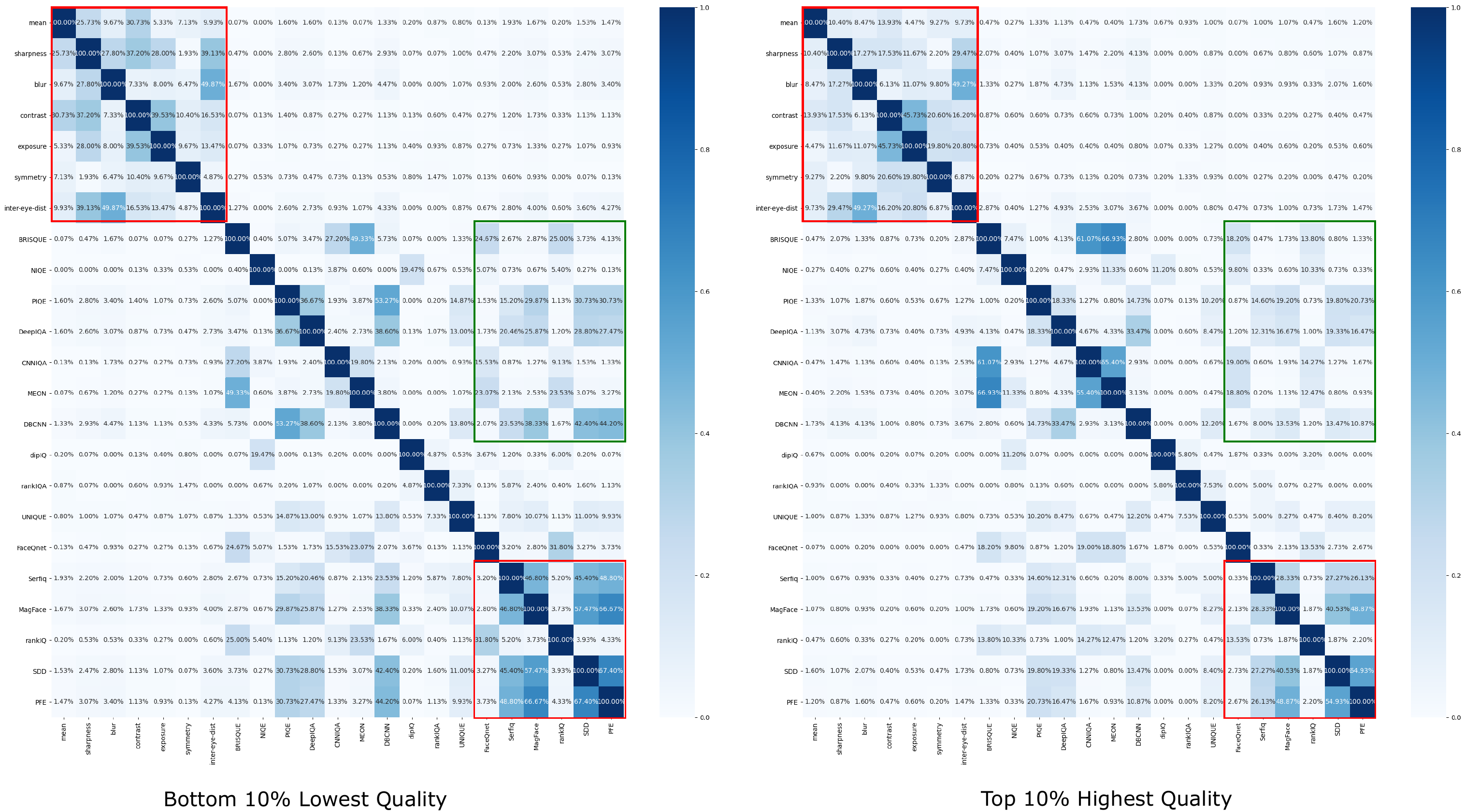}
    \caption{\bfu{The confusion matrix shows the
    ratio of overlapped samples between the samples with the lowest/highest 10\% qualities (lowest on the left matrix and highest on the right matrix) as measured by two quality estimation methods (on the X and Y axes). A high intra-category overlap is seen in the handcrafted-based FIQA, as well as, the DL-based FIQA, both in red squares. A relatively high inter-category correlation is noticed between IQA and DL-based FIQA methods (in green box). This matrix is build on VGGFace2. Additional results for other two databases are provided in the supplementary materials.}
    }
    \label{fig:confusion_matrix}
\end{figure*}

\bfu{Even though most handcrafted features show high intra-category correlation, e.g., contrast and exposure with an overlapping ratio of 39.53\%, or contrast and mean with 30.73\%, according to Figure \ref{fig:confusion_matrix}, the inter-category correlation is relative low for these metrics. In general, individual features \cite{ISOIEC29794-5} show low impact on selecting high-utility face images. This is confirmed in Figure \ref{fig:confusion_matrix} (left) where an average ratio of less than 1.5\% is shown between the FIQA handcrafted features and DL-based FIQA methods.}
\bfu{To sum up, these handcrafted features are sensitive to the image capture setups and the post-processing steps performed on the captured images. Therefore, these features might be less useful in comparison with learned FIQA methods.}

\paragraph{Q2: How does IQA metrics correlate to face image utility?}

\begin{figure*}
\centering
\includegraphics[width=.95\linewidth]{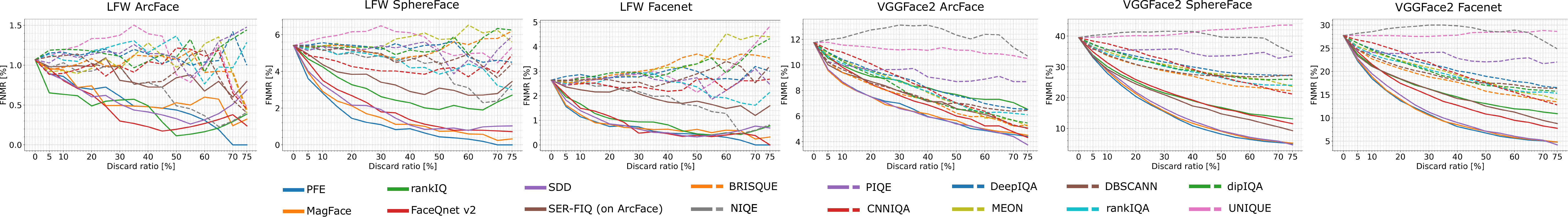}
\caption{Error vs.~reject characteristics at FMR1000 for IQA (dashed lines) and FIQA (solid lines) methods. The rows show the results for diverse face embeddings on LFW and VGGFace2. DL-based FIQA methods outperform IQA methods, while for VGGFace2 a clear decreasing trend can be observed for IQA methods as well.} 
\label{fig:erc_001_net}
\end{figure*}

Figure~\ref{fig:erc_001_net} depicts the ERCs at FMR1000 using the DL-based FIQA methods from Section~\ref{methods_fiqa} (solid lines) and learning-based IQA methods from Section~\ref{methods_iqa} (dashed lines). Taking a look at the results on LFW (the left three plots in Figure \ref{fig:erc_001_net}), we observed that error is reduced due to the increasing amount of dropped low-quality images. In contrast to the findings on LFW, ERC on VGGFace2 (the right three plots in Figure \ref{fig:erc_001_net}) exhibits a strong correlation between the learned IQA methods towards image utility. The curves using IQA methods reveal a distinctly decreasing FNMR when discarding bad quality images. The total reduction in error for learning-based IQA methods is more evident compared to FIQA methods or handcrafted quality metrics. Observing the ERC curve for VGGFace2 at FMR1000 with ArcFace embeddings, the error in terms of FNMR for CNNIQA dropped around 36.7\% from 20\% to 40\% reject ratios, whereas FaceQnet only reduced about 19\% and the inter-eye-distance feature about 16\%.

\bfu{Table~\ref{tab:top_perform} showed that although FIQA metrics outperform the most other metrics according to the verification results. Nevertheless, in most settings, the learned IQA methods can also compete for the top-3 rank. As on VGGFace2, they all show a strong correlation with a clear trend towards decreasing error rates, with no clear winner in this category. The same result is confirmed in Figure \ref{fig:confusion_matrix}. Besides the intra-category correlation, we also observed high overlapping ratios between the IQA metrics and the DL-based FIQA metrics, unlike handcrafted features and DL-based FIQAs. Especially IQA methods, such as PIQE, DeepIQA and DBCNN showed large overlap to MagFace, SDD-FIQA, and PFE, where the highest overlap is observed for DBCNN to PFE (44.20\%). These three methods are representative of the model-based, CNN-based, and multitask-based IQA approaches, while the ranking-based approaches showed less correlation to DL-based FIQAs.}

\begin{figure*}
    \centering
    \includegraphics[width=.95\linewidth]{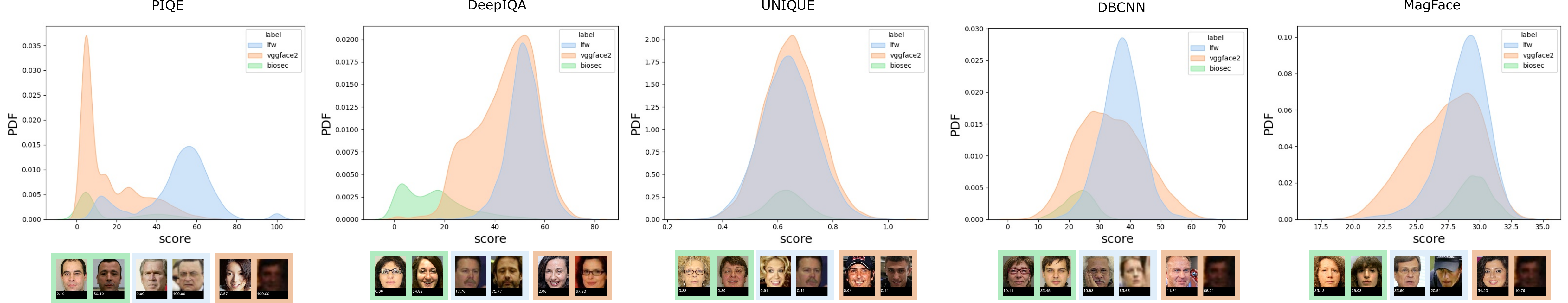}
    \caption{\bfu{Score distribution for the three face database using one representative method from 4 IQA sub-categories and 1 FIQA method for reference. For each method and database, one high utility sample (left) and one low utility sample (right) are displayed. VGGFace2 has a wider score distribution compared to LFW and Biosecure in most cases, which might indicate that the IQA methods are more effective on this database.}}
    \label{fig:visu_samples}
\end{figure*}

\bfu{Looking at the score distributions in Figure \ref{fig:visu_samples}, we observed that VGGFace2 database has a wider score distribution compared to LFW and Biosecure. This diversity can be the reason behind effectiveness of IQA on VGGFace2 database compared to the other databases.}

\bfu{To conclude, there is a clear correlation between IQA and face image utility. The drop in the ERC curve shows the effectiveness of the IQA metrics with respect to select the face image with high utility. This drop is relatively stronger for model-based and multitask-based IQA approaches.}

\paragraph{Q3: How does DL-based FIQA metrics correlate to face image utility?}

Considering the verification results in Table~\ref{tab:top_perform}, in general, one of the DL-based FIQA methods always dominates the top verification results independent of the database. Inspecting the results for VGGFace2 at FMR1000, SDD-FIQA and MagFace are the methods that have occupied the top rankings most of the times. One explanation for the good outcomes in terms of error performance is that FIQA methods are trained with face images for specific FR systems as in the proposed methods \cite{DBLP:journals/corr/abs-2103-05977, meng2021magface}.

\bfu{Typical FIQA methods like SDD-FIQA, MagFace, SER-FIQ, and  PFE, already learned discriminative face embeddings in the training phase using appropriate loss functions. Other FIQA methods like rankIQ and FaceQnet learned the utility directly from relative FR comparison scores. Moreover, face image utility is not fully represented by the perceptual quality and is not easily quantifiable. Therefore, opposite to general IQA metrics, FIQA inherently learn the face utility. In Figure \ref{fig:confusion_matrix}, most DL-based FIQA methods show a high intra-category correlation. Especially the unsupervised methods, such as SER-FIQ, MagFace, SDD-FIQA, and PFE revealed a strong intra-category correlation, while FaceQnet and rankIQ possess low correlation to other DL-based FIQA metrics. The highest overlap is observed for SDD-FIQA to PFE with an overlap of 67.40\% and MagFace to PFE with 66.67\%.}

\bfu{To summarize, FIQA metrics are designed to relate a face image to face utility. Even trained for certain face embeddings, most FIQA methods work efficiently across different FR solutions. Unsupervised methods, such as SDD-FIQA, PFE, and MagFace outperform other supervised methods, like FaceQnet.}

\begin{table*}
\resizebox{\textwidth}{!}{%
\begin{tabular}{lllllllllllll}\hline
\multicolumn{13}{c}{\textbf{LFW at FMR 1000}} \\ \hline
\multicolumn{1}{l|}{} &
  \multicolumn{4}{c|}{ArcFace} &
  \multicolumn{4}{c|}{Sphereface} &
  \multicolumn{4}{c|}{FaceNet} \\ \cline{2-13} 
\multicolumn{1}{l|}{} &
  \multicolumn{2}{c|}{20\%} &
  \multicolumn{2}{c|}{40\%} &
  \multicolumn{2}{c|}{20\%} &
  \multicolumn{2}{c|}{40\%} &
  \multicolumn{2}{c|}{20\%} &
  \multicolumn{2}{c|}{40\%} \\ \hline
\multicolumn{1}{l|}{DL-based FIQA} &
  \multicolumn{1}{l|}{\textbf{rankIQ}} &
  \multicolumn{1}{l|}{\textbf{0.488}} &
  \multicolumn{1}{l|}{\textbf{FaceQnet}} &
  \multicolumn{1}{l|}{\textbf{0.226}} &
  \multicolumn{1}{l|}{\textbf{PFE}} &
  \multicolumn{1}{l|}{\textbf{1.444}} &
  \multicolumn{1}{l|}{\textbf{PFE}} &
  \multicolumn{1}{l|}{\textbf{0.877}} &
  \multicolumn{1}{l|}{\textbf{PFE}} &
  \multicolumn{1}{l|}{\textbf{0.747}} &
  \multicolumn{1}{l|}{\textbf{FaceQnet}} &
  \multicolumn{1}{l|}{\textbf{0.453}}\\ \cline{2-13} 
\multicolumn{1}{l|}{} &
  \multicolumn{1}{l|}{\textbf{SDD}} &
  \multicolumn{1}{l|}{\textbf{0.602}} &
  \multicolumn{1}{l|}{\textbf{SDD}} &
  \multicolumn{1}{l|}{\textbf{0.412}} &
  \multicolumn{1}{l|}{\textbf{MagFace}} &
  \multicolumn{1}{l|}{\textbf{2.125}} &
  \multicolumn{1}{l|}{\textbf{SDD}} &
  \multicolumn{1}{l|}{\textbf{1.072}} &
  \multicolumn{1}{l|}{\textbf{MagFace}} &
  \multicolumn{1}{l|}{\textbf{0.840}} &
  \multicolumn{1}{l|}{\textbf{PFE}} &
  \multicolumn{1}{l|}{\textbf{0.478}} \\ \cline{2-13} 
\multicolumn{1}{l|}{} &
  \multicolumn{1}{l|}{FaceQnet} &
  \multicolumn{1}{l|}{0.632} &
  \multicolumn{1}{l|}{\textbf{PFE}} &
  \multicolumn{1}{l|}{\textbf{0.478}} &
  \multicolumn{1}{l|}{\textbf{SDD}} &
  \multicolumn{1}{l|}{\textbf{2.159}} &
  \multicolumn{1}{l|}{\textbf{MagFace}} &
  \multicolumn{1}{l|}{\textbf{1.118}} &
  \multicolumn{1}{l|}{\textbf{SDD}} &
  \multicolumn{1}{l|}{\textbf{0.853}} &
  \multicolumn{1}{l|}{\textbf{SDD}} &
  \multicolumn{1}{l|}{\textbf{0.495}} \\ \hline
\multicolumn{1}{l|}{Image Quality} &
  \multicolumn{1}{l|}{NIQE} &
  \multicolumn{1}{l|}{0.925} &
  \multicolumn{1}{l|}{NIQE} &
  \multicolumn{1}{l|}{0.780} &
  \multicolumn{1}{l|}{CNNIQA} &
  \multicolumn{1}{l|}{4.263} &
  \multicolumn{1}{l|}{CNNIQA} &
  \multicolumn{1}{l|}{3.930} &
  \multicolumn{1}{l|}{CNNIQA} &
  \multicolumn{1}{l|}{2.385} &
  \multicolumn{1}{l|}{NIQE} &
  \multicolumn{1}{l|}{1.994} \\ \cline{2-13} 
\multicolumn{1}{l|}{} &
  \multicolumn{1}{l|}{MEON} &
  \multicolumn{1}{l|}{0.959} &
  \multicolumn{1}{l|}{DBCNN} &
  \multicolumn{1}{l|}{0.786} &
  \multicolumn{1}{l|}{NIQE} &
  \multicolumn{1}{l|}{4.989} &
  \multicolumn{1}{l|}{rankIQA} &
  \multicolumn{1}{l|}{4.188} &
  \multicolumn{1}{l|}{NIQE} &
  \multicolumn{1}{l|}{2.417} &
  \multicolumn{1}{l|}{CNNIQA} &
  \multicolumn{1}{l|}{2.183} \\ \cline{2-13} 
\multicolumn{1}{l|}{} &
  \multicolumn{1}{l|}{CNNIQA} &
  \multicolumn{1}{l|}{1.015} &
  \multicolumn{1}{l|}{CNNIQA} &
  \multicolumn{1}{l|}{1.048} &
  \multicolumn{1}{l|}{rankIQA} &
  \multicolumn{1}{l|}{5.071} &
  \multicolumn{1}{l|}{DBCNN} &
  \multicolumn{1}{l|}{4.628} &
  \multicolumn{1}{l|}{DBCNN} &
  \multicolumn{1}{l|}{2.651} &
  \multicolumn{1}{l|}{DBCNN} &
  \multicolumn{1}{l|}{2.532} \\ \hline
\multicolumn{1}{l|}{Feature FIQA} &
  \multicolumn{1}{l|}{inter eye dist} &
  \multicolumn{1}{l|}{.9027} &
  \multicolumn{1}{l|}{inter eye dist} &
  \multicolumn{1}{l|}{.8319} &
  \multicolumn{1}{l|}{inter eye dist} &
  \multicolumn{1}{l|}{4.9147} &
  \multicolumn{1}{l|}{inter eye dist} &
  \multicolumn{1}{l|}{4.4925} &
  \multicolumn{1}{l|}{inter eye dist} &
  \multicolumn{1}{l|}{2.6579} &
  \multicolumn{1}{l|}{inter eye dist} &
  \multicolumn{1}{l|}{2.7454} \\ \cline{2-13} 
\multicolumn{1}{l|}{} &
  \multicolumn{1}{l|}{\textbf{mean}} &
  \multicolumn{1}{l|}{\textbf{.5326}} &
  \multicolumn{1}{l|}{mean} &
  \multicolumn{1}{l|}{.5586} &
  \multicolumn{1}{l|}{mean} &
  \multicolumn{1}{l|}{3.7288} &
  \multicolumn{1}{l|}{mean} &
  \multicolumn{1}{l|}{3.2721} &
  \multicolumn{1}{l|}{mean} &
  \multicolumn{1}{c|}{1.7917} &
  \multicolumn{1}{l|}{mean} &
  \multicolumn{1}{c|}{1.9154} \\ \cline{2-13} 
\multicolumn{1}{l|}{} &
  \multicolumn{1}{l|}{sharpness} &
  \multicolumn{1}{l|}{1.2487} &
  \multicolumn{1}{l|}{sharpness} &
  \multicolumn{1}{l|}{1.3559} &
  \multicolumn{1}{l|}{sharpness} &
  \multicolumn{1}{l|}{5.7942} &
  \multicolumn{1}{l|}{sharpness} &
  \multicolumn{1}{l|}{6.2711} &
  \multicolumn{1}{l|}{sharpness} &
  \multicolumn{1}{l|}{2.7972} &
  \multicolumn{1}{l|}{sharpness} &
  \multicolumn{1}{l|}{3.2203} \\ \hline
\\
\hline
\multicolumn{13}{c}{\textbf{VGGFace2 at FMR 1000}} \\ \hline
\multicolumn{1}{l|}{} &
  \multicolumn{4}{c|}{ArcFace} &
  \multicolumn{4}{c|}{Sphereface} &
  \multicolumn{4}{c|}{FaceNet} \\ \cline{2-13} 
\multicolumn{1}{l|}{} &
  \multicolumn{2}{c|}{20\%} &
  \multicolumn{2}{c|}{40\%} &
  \multicolumn{2}{c|}{20\%} &
  \multicolumn{2}{c|}{40\%} &
  \multicolumn{2}{c|}{20\%} &
  \multicolumn{2}{c|}{40\%} \\ \hline
\multicolumn{1}{l|}{DL-based FIQA} &
  \multicolumn{1}{l|}{\textbf{PFE}} &
  \multicolumn{1}{l|}{\textbf{7.505}} &
  \multicolumn{1}{l|}{\textbf{SDD}} &
  \multicolumn{1}{l|}{\textbf{5.931}} &
  \multicolumn{1}{l|}{\textbf{PFE}} &
  \multicolumn{1}{l|}{\textbf{20.394}} &
  \multicolumn{1}{l|}{\textbf{PFE}} &
  \multicolumn{1}{l|}{\textbf{10.836}} &
  \multicolumn{1}{l|}{\textbf{PFE}} &
  \multicolumn{1}{l|}{\textbf{13.694}} &
  \multicolumn{1}{l|}{\textbf{PFE}} &
  \multicolumn{1}{l|}{\textbf{8.127}} \\ \cline{2-13} 
\multicolumn{1}{l|}{} &
  \multicolumn{1}{l|}{\textbf{SDD}} &
  \multicolumn{1}{l|}{\textbf{7.508}} &
  \multicolumn{1}{l|}{\textbf{PFE}} &
  \multicolumn{1}{l|}{\textbf{5.955}} &
  \multicolumn{1}{l|}{\textbf{MagFace}} &
  \multicolumn{1}{l|}{\textbf{21.329}} &
  \multicolumn{1}{l|}{\textbf{MagFace}} &
  \multicolumn{1}{l|}{\textbf{11.650}} &
  \multicolumn{1}{l|}{\textbf{MagFace}} &
  \multicolumn{1}{l|}{\textbf{13.993}} &
  \multicolumn{1}{l|}{\textbf{MagFace}} &
  \multicolumn{1}{l|}{\textbf{8.540}} \\ \cline{2-13} 
\multicolumn{1}{l|}{} &
  \multicolumn{1}{l|}{\textbf{MagFace}} &
  \multicolumn{1}{l|}{\textbf{7.520}} &
  \multicolumn{1}{l|}{\textbf{MagFace}} &
  \multicolumn{1}{l|}{\textbf{6.171}} &
  \multicolumn{1}{l|}{\textbf{SDD}} &
  \multicolumn{1}{l|}{\textbf{21.836}} &
  \multicolumn{1}{l|}{\textbf{SDD}} &
  \multicolumn{1}{l|}{\textbf{12.371}} &
  \multicolumn{1}{l|}{\textbf{SDD}} &
  \multicolumn{1}{l|}{\textbf{14.905}} &
  \multicolumn{1}{l|}{\textbf{SDD}} &
  \multicolumn{1}{l|}{\textbf{8.987}} \\ \hline
\multicolumn{1}{l|}{Image Quality} &
  \multicolumn{1}{l|}{BRISQUE} &
  \multicolumn{1}{l|}{8.468} &
  \multicolumn{1}{l|}{dipIQ} &
  \multicolumn{1}{l|}{7.234} &
  \multicolumn{1}{l|}{BRISQUE} &
  \multicolumn{1}{l|}{30.706} &
  \multicolumn{1}{l|}{BRISQUE} &
  \multicolumn{1}{l|}{27.002} &
  \multicolumn{1}{l|}{BRISQUE} &
  \multicolumn{1}{l|}{20.593} &
  \multicolumn{1}{l|}{BRISQUE} &
  \multicolumn{1}{l|}{17.683} \\ \cline{2-13} 
\multicolumn{1}{l|}{} &
  \multicolumn{1}{l|}{DBCNN} &
  \multicolumn{1}{l|}{8.565} &
  \multicolumn{1}{l|}{DBCNN} &
  \multicolumn{1}{l|}{7.246} &
  \multicolumn{1}{l|}{DBCNN} &
  \multicolumn{1}{l|}{30.813} &
  \multicolumn{1}{l|}{DBCNN} &
  \multicolumn{1}{l|}{27.502} &
  \multicolumn{1}{l|}{DBCNN} &
  \multicolumn{1}{l|}{21.012} &
  \multicolumn{1}{l|}{DBCNN} &
  \multicolumn{1}{l|}{18.144} \\ \cline{2-13} 
\multicolumn{1}{l|}{} &
  \multicolumn{1}{l|}{dipIQ} &
  \multicolumn{1}{l|}{9.248} &
  \multicolumn{1}{l|}{BRISQUE} &
  \multicolumn{1}{l|}{7.292} &
  \multicolumn{1}{l|}{rankIQA} &
  \multicolumn{1}{l|}{32.459} &
  \multicolumn{1}{l|}{rankIQA} &
  \multicolumn{1}{l|}{28.355} &
  \multicolumn{1}{l|}{rankIQA} &
  \multicolumn{1}{l|}{22.066} &
  \multicolumn{1}{l|}{dipIQ} &
  \multicolumn{1}{l|}{18.519} \\ \hline
\multicolumn{1}{l|}{Feature FIQA} &
  \multicolumn{1}{l|}{inter eye dist} &
  \multicolumn{1}{l|}{8.669} &
  \multicolumn{1}{l|}{inter eye dist} &
  \multicolumn{1}{l|}{7.2749} &
  \multicolumn{1}{l|}{inter eye dist} &
  \multicolumn{1}{l|}{28.378} &
  \multicolumn{1}{l|}{inter eye dist} &
  \multicolumn{1}{l|}{25.373} &
  \multicolumn{1}{l|}{inter eye dist} &
  \multicolumn{1}{l|}{19.364} &
  \multicolumn{1}{l|}{inter eye dist} &
  \multicolumn{1}{l|}{16.701} \\ \cline{2-13} 
\multicolumn{1}{l|}{} &
  \multicolumn{1}{l|}{mean} &
  \multicolumn{1}{l|}{10.611} &
  \multicolumn{1}{l|}{mean} &
  \multicolumn{1}{l|}{9.7969} &
  \multicolumn{1}{l|}{mean} &
  \multicolumn{1}{l|}{36.088} &
  \multicolumn{1}{l|}{mean} &
  \multicolumn{1}{l|}{34.573} &
  \multicolumn{1}{l|}{mean} &
  \multicolumn{1}{c|}{25.628} &
  \multicolumn{1}{l|}{mean} &
  \multicolumn{1}{c|}{24.578} \\ \cline{2-13} 
\multicolumn{1}{l|}{} &
  \multicolumn{1}{l|}{sharpness} &
  \multicolumn{1}{l|}{9.373} &
  \multicolumn{1}{l|}{sharpness} &
  \multicolumn{1}{l|}{8.3267} &
  \multicolumn{1}{l|}{sharpness} &
  \multicolumn{1}{l|}{30.967} &
  \multicolumn{1}{l|}{sharpness} &
  \multicolumn{1}{l|}{27.952} &
  \multicolumn{1}{l|}{sharpness} &
  \multicolumn{1}{l|}{21.145} &
  \multicolumn{1}{l|}{sharpness} &
  \multicolumn{1}{l|}{18.704} \\ \hline
\\
\end{tabular}%
}
\caption{Comparison between the top-3 performing DL-based FIQA, feature-based metrics, and IQA methods on two facial image DBs evaluated for the FNMR at two reject ratios (20\,\% and 40\,\%) at FMR1000 based on three FR models (ArcFace, SphereFace, Facenet). In bold are the best performing three methods across all categories.}
\label{tab:top_perform}
\end{table*}

\section{Visualization of selected model decisions}
\label{sec:visualization} 

\begin{figure}
\includegraphics[width=.99\columnwidth]{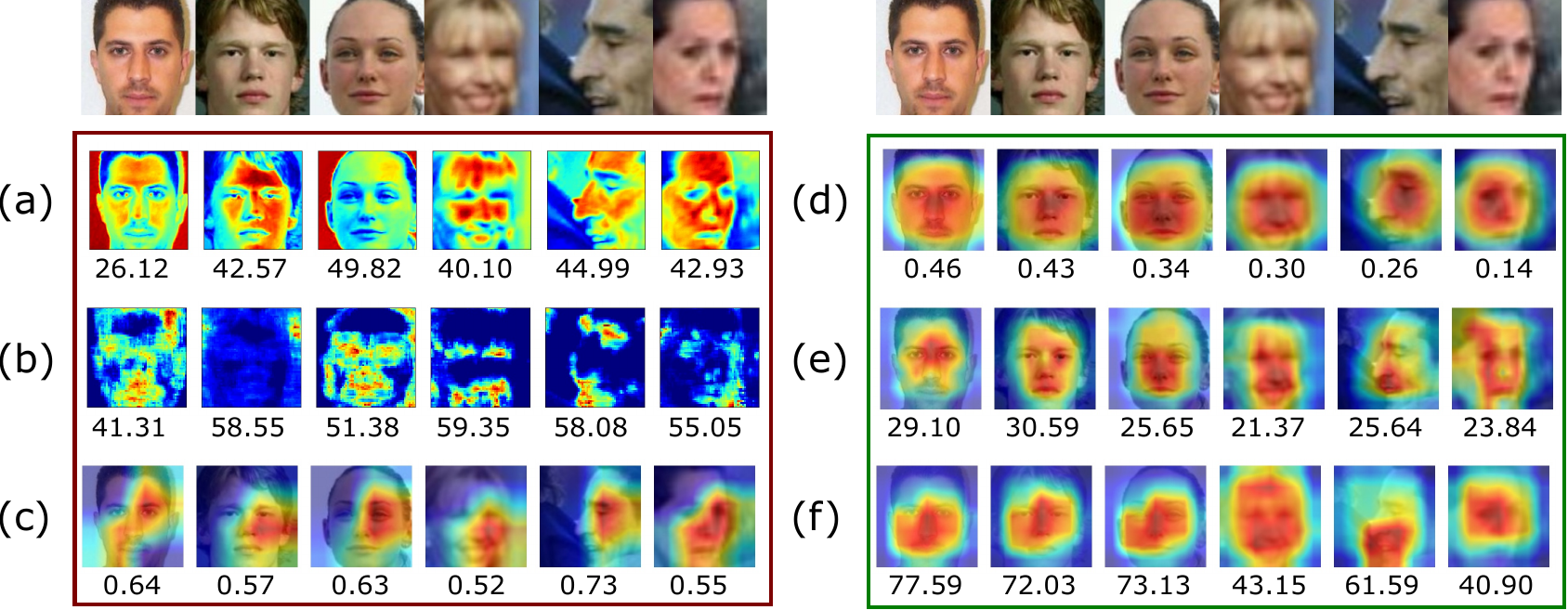}
\caption{\bfu{visualizes images of diverse network's attention and face quality score. Left are the three IQA methods: (a) CNNIQA, (b) DeepIQA, and (c) UNIQUE, and right are the three FIQA methods: (d) FaceQnet, (e) MagFace, and (f) SDD-FIQA. The IQA methods focus their decision additionally on the background, while FIQA methods focus mainly on center face area.}}
\label{fig:method_viz}
\end{figure}

In addition, to evaluate the effectiveness of the different investigated methods in measuring face image utility, we go further by exploring the image parts that contribute the most to their performance, especially those from IQA and FIQA categories. \bfu{Figure~\ref{fig:method_viz} visualizes three selected IQA methods (left) and three FIQA methods (right). We use the Score-CAM \cite{wang2020score} to display the attention map of the network's decision. Score-CAM is designed to display visual explanations for CNNs and works more efficient compared to other gradient-based visualization methods. UNIQUE, FaceQnet, MagFace, and SDD-FIQA are visualized using their ResNet-X base architecture. This visualization is unfortunately not applicable to other methods, such as patch-based approaches, or rankIQ due to the fusion of selected features in a two-stage process, and the SER-FIQ due to multiple dropout runs to modify network architectures.}

\textit{Visualization for IQA methods: }
For patch-based IQA, such as CNNIQA and DeepIQA, pixel-wise network decision is visualized in Figure~\ref{fig:method_viz}. It is observed that these IQA methods also paid attention to the background (see first and third images in Figure \ref{fig:method_viz}(a)) for CNNIQA. Such behavior might be the reason for the lower performance. Similar results are seen for DeepIQA (Figure~\ref{fig:method_viz}(b)). \bfu{Score-CAM is used for the ranking-based IQA method UNIQUE. Consistent finding with the other two IQA methods are observed for the third and sixth image in Figure \ref{fig:method_viz}(c). Opposite to DL-based FIQA methods, the networks attention for IQA methods focuses not fully on the center face area.}

\textit{Visualization for FIQA methods: }
\bfu{Looking at Figure~\ref{fig:method_viz} (d) to (f), the activation area for FIQA methods cover mostly the central part of a face. From Figure \ref{fig:method_viz}(a) to (c), it is already noticed that the IQA methods have also relatively higher contributions on non-facial areas of the image. In contrast, the FIQA methods mainly focus on the facial areas and neglect the background. This may be one of the main reasons behind the superior performance of such methods. Furthermore, it is to note that SDD-FIQA and MagFace have more locally refined attention compared to FaceQnet.}

\section{Conclusions}
\label{conclusion}

To address research questions concerning the face image utility correlation with general image quality and face-specific quality metrics, we thoroughly investigated a total of \bfu{25 quality metrics}. We divided these into four different categories in 1) general image quality measures, 2) handcrafted image quality measures, 3) face-related handcrafted measures, and 4) learned face utility measures. These quality metrics are evaluated on three databases and three FR solutions and provided a deep discussion on the relationship between these families of quality estimation solutions.

Our evaluation showed little influence of single handcrafted feature as face image utility. \bfu{Although these features revealed a clear intra-category correlation, they show less inter-category correlation to other learned IQA or FIQA metrics. Therefore, they demonstrate no clear indications to be useful as a generalized metric to assess face image utility.}
DL-based FIQA methods, as a face image utility predictor, are optimized for face images and show superior performance over IQA methods across different setups. Nevertheless, IQA methods show a clear correlation to face image utility, even though they do not outperform DL-based FIQA methods. Visualization of the IQA methods output revealed a focus on areas in the background, rather than solely on the face as in the learned FIQA methods. Accompanied with the advantage of FR model-independent training of such IQA methods, combining IQA metric with DL-based FIQA metric could lead to a more generalized measure across different FR systems and application scenarios.

\textbf{Acknowledgements:}
This research work has been funded by the German Federal Ministry of Education and Research and the Hessian Ministry of Higher Education, Research, Science and the Arts within their joint support of the National Research Center for Applied Cybersecurity ATHENE.

\begin{figure*}
    \centering
    \includegraphics[page=1, width=\linewidth]{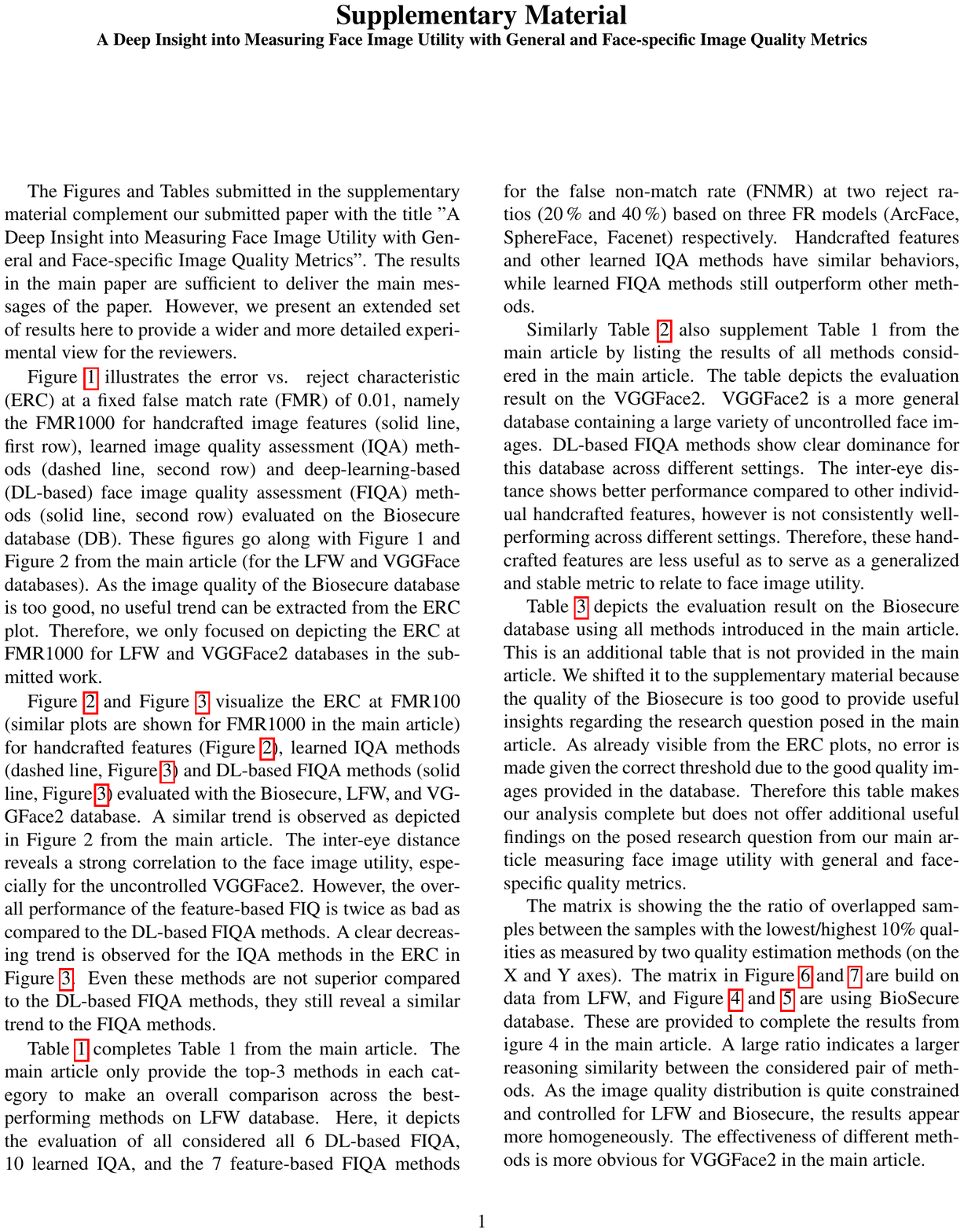}
\end{figure*}

\begin{figure*}
    \centering
    \includegraphics[page=2, width=\linewidth]{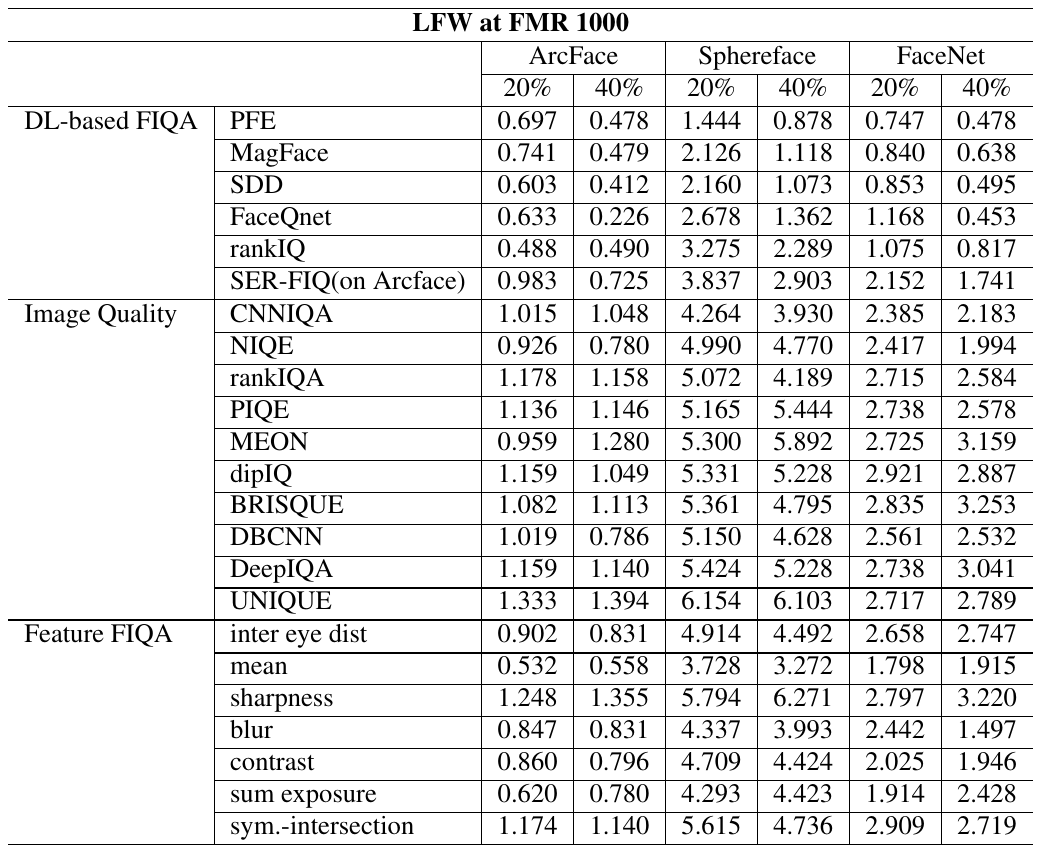}
\end{figure*}

\begin{figure*}
    \centering
    \includegraphics[page=3, width=\linewidth]{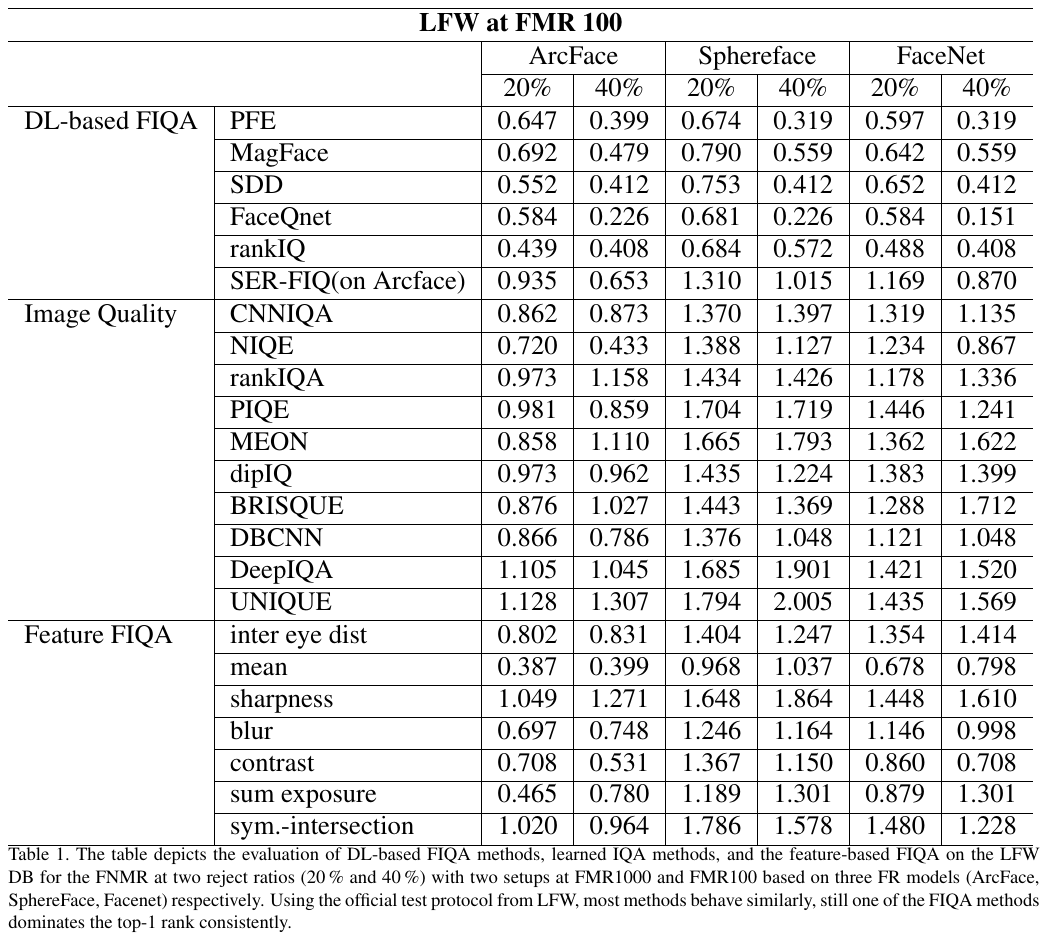}
\end{figure*}

\begin{figure*}
    \centering
    \includegraphics[page=4, width=\linewidth]{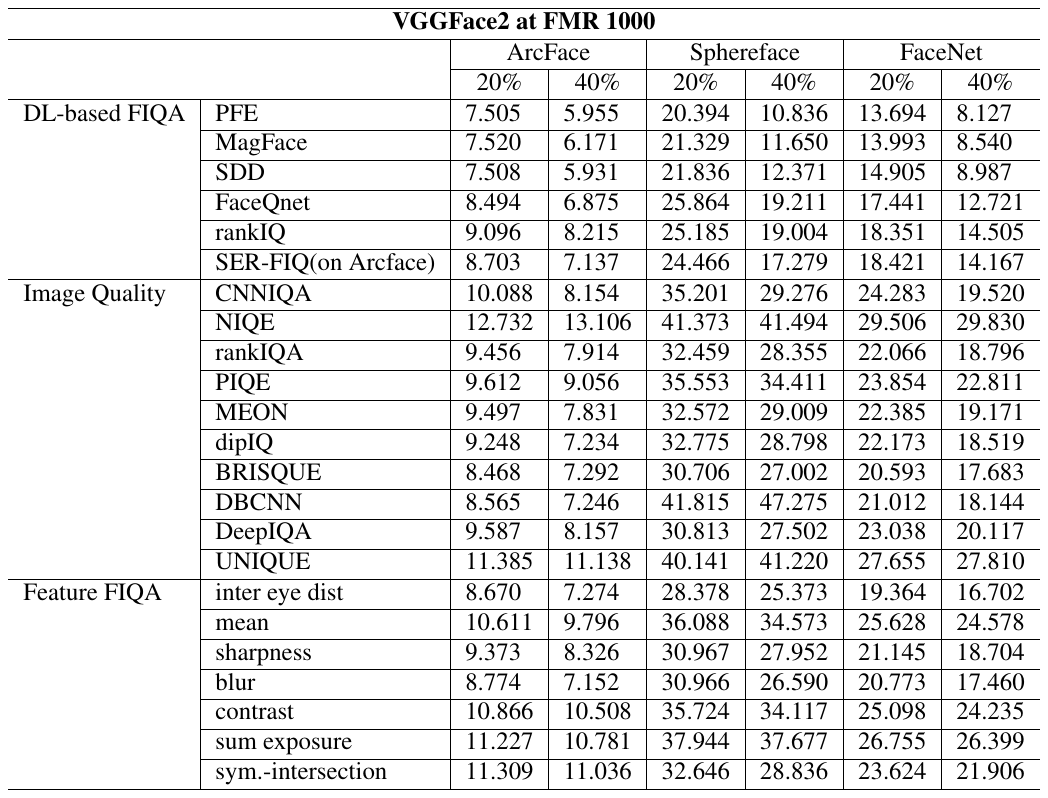}
\end{figure*}

\begin{figure*}
    \centering
    \includegraphics[page=5, width=\linewidth]{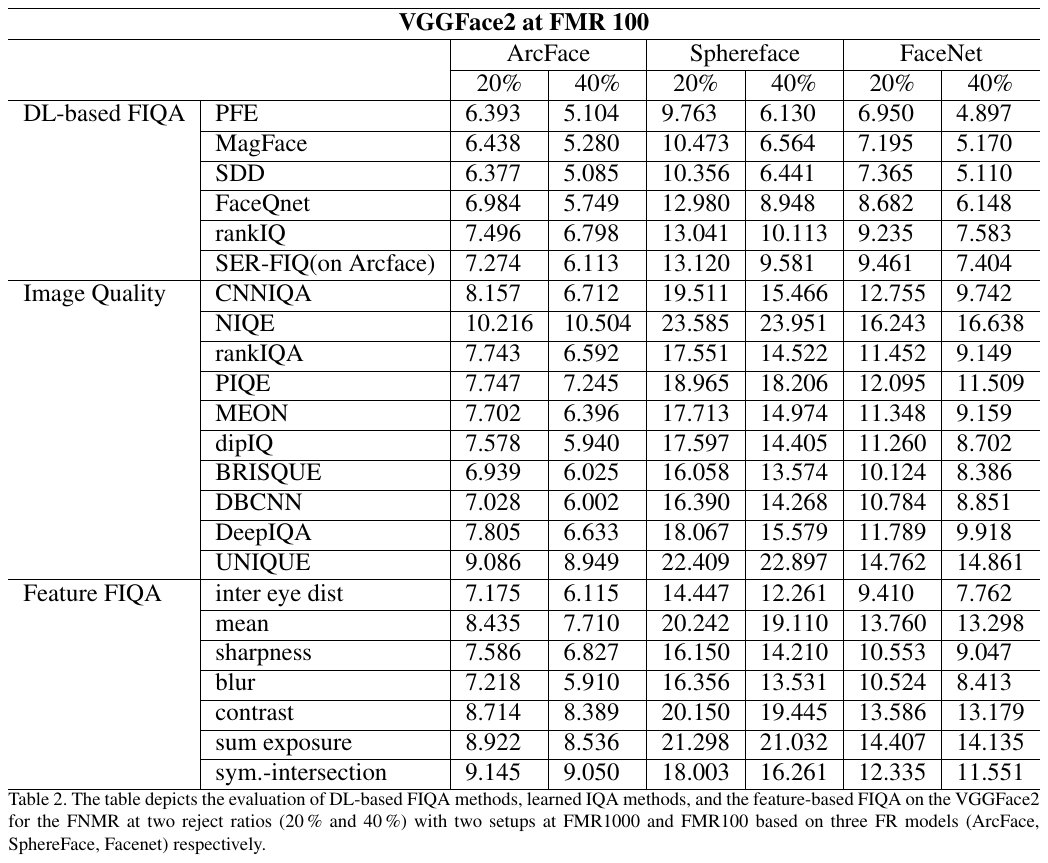}
\end{figure*}

\begin{figure*}
    \centering
    \includegraphics[page=6, width=\linewidth]{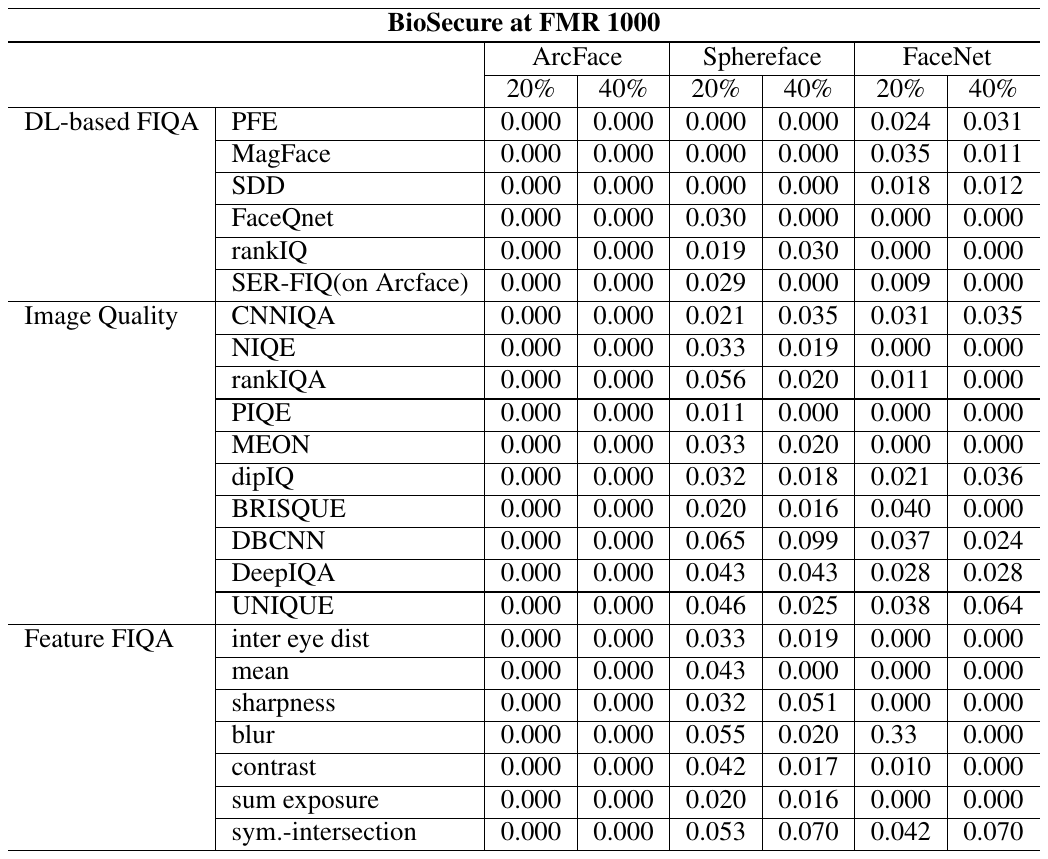}
\end{figure*}

\begin{figure*}
    \centering
    \includegraphics[page=7, width=\linewidth]{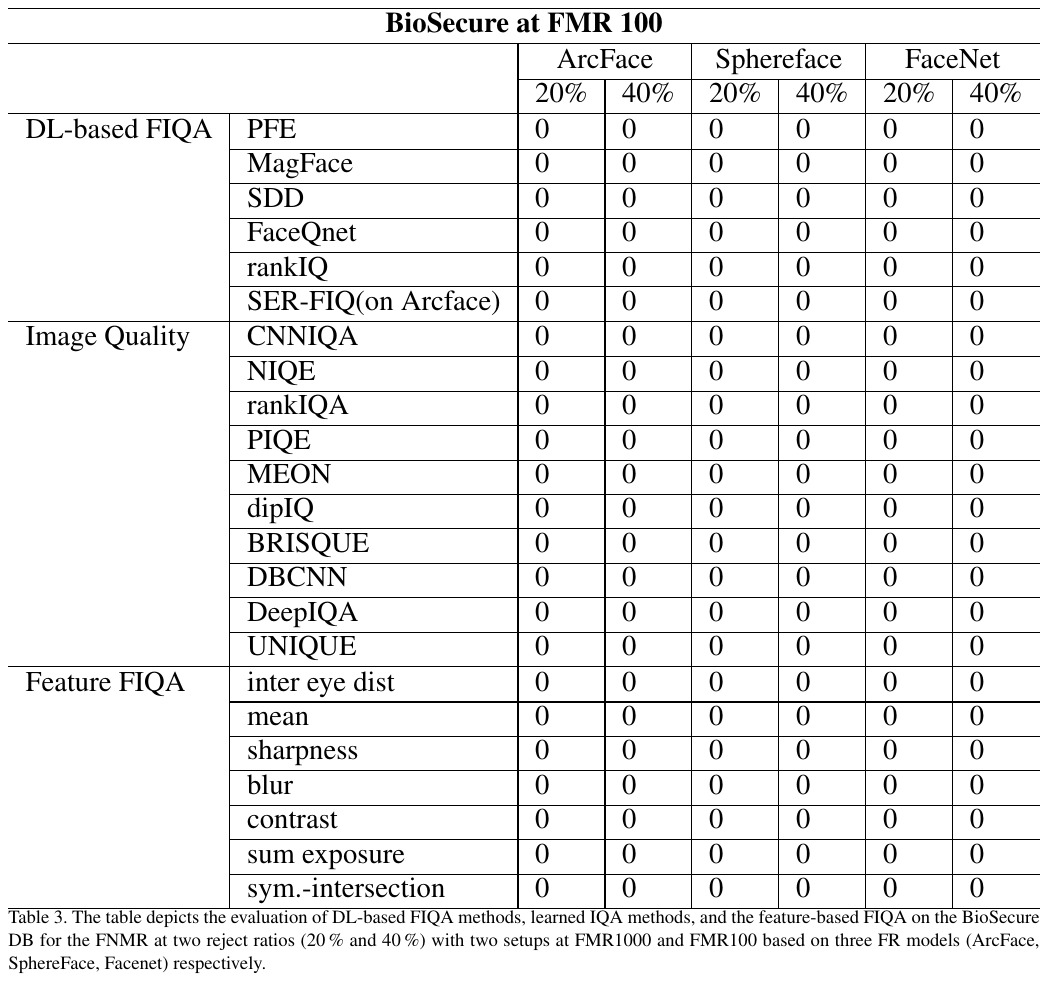}
\end{figure*}

\begin{figure*}
    \centering
    \includegraphics[page=8, width=\linewidth]{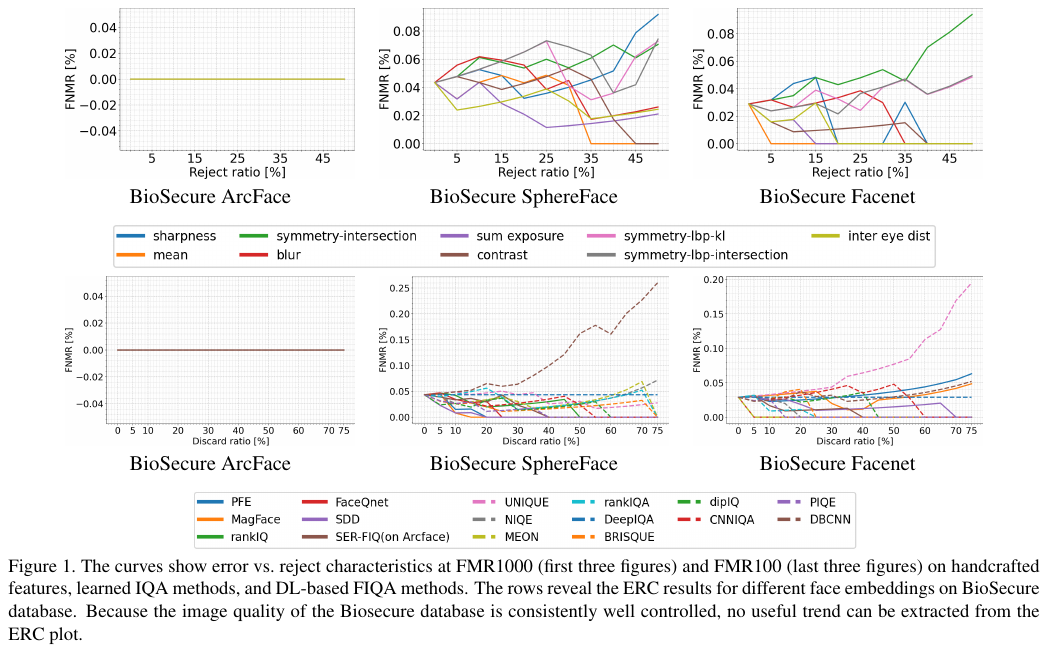}
\end{figure*}

\begin{figure*}
    \centering
    \includegraphics[page=9, width=\linewidth]{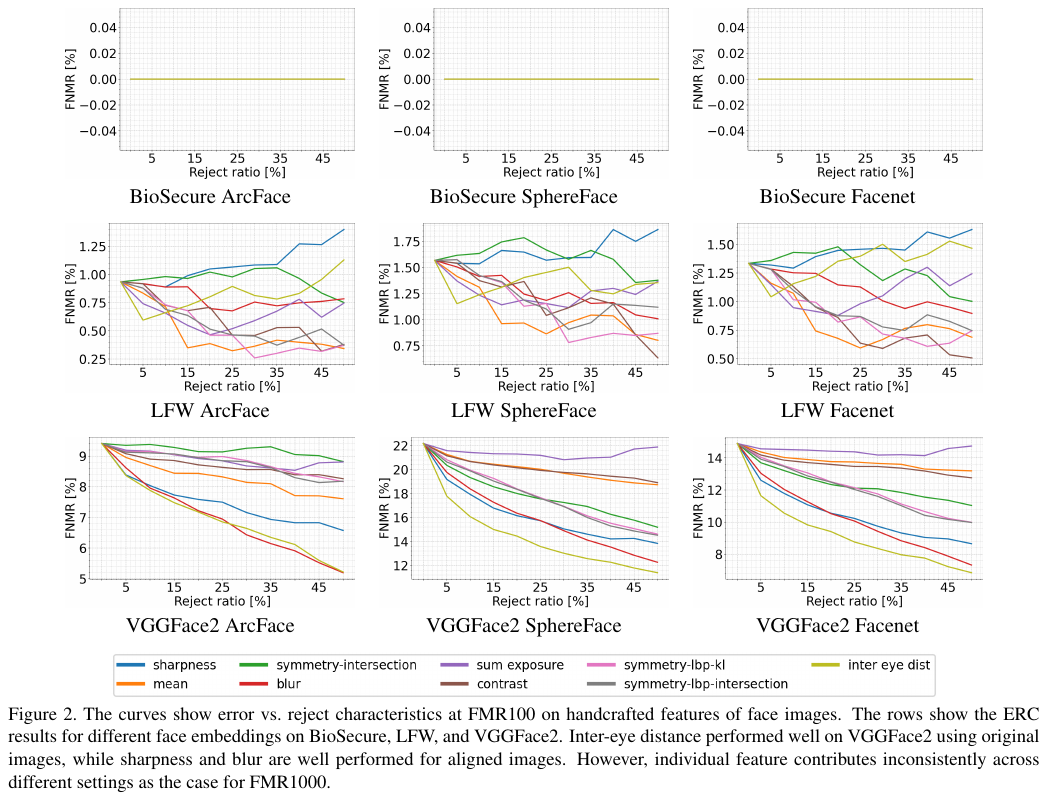}
\end{figure*}

\begin{figure*}
    \centering
    \includegraphics[page=10, width=\linewidth]{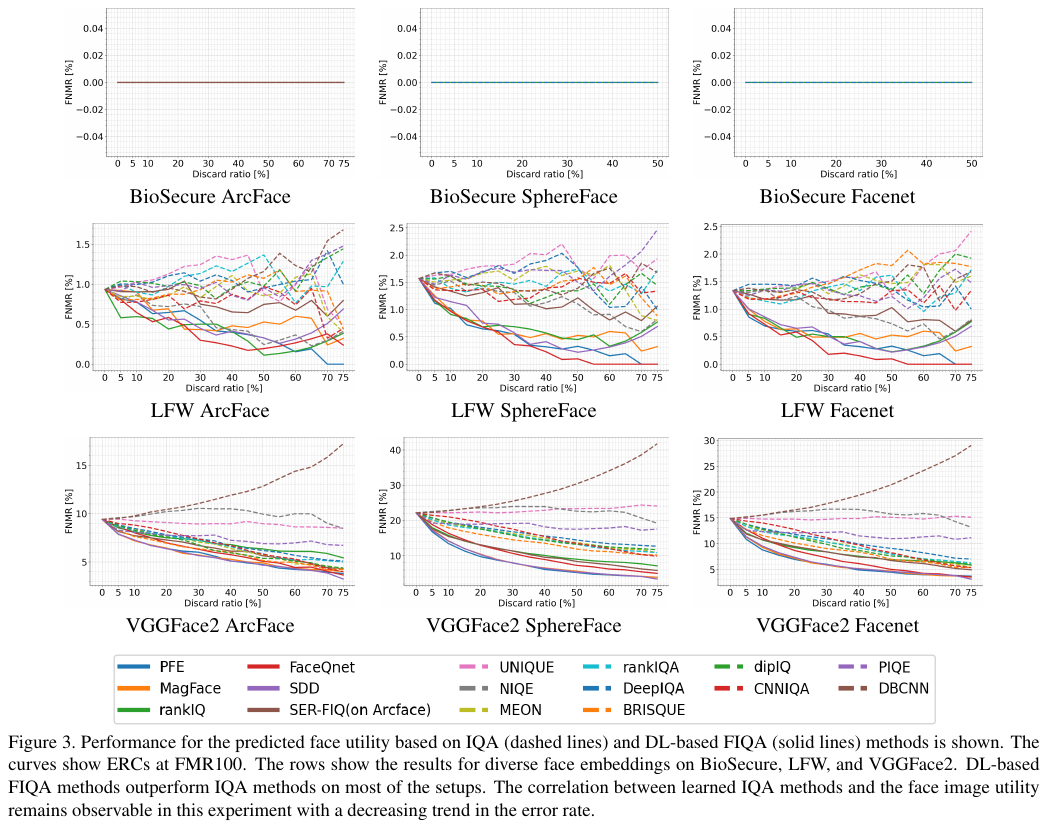}
\end{figure*}

\begin{figure*}
    \centering
    \includegraphics[page=11, width=\linewidth]{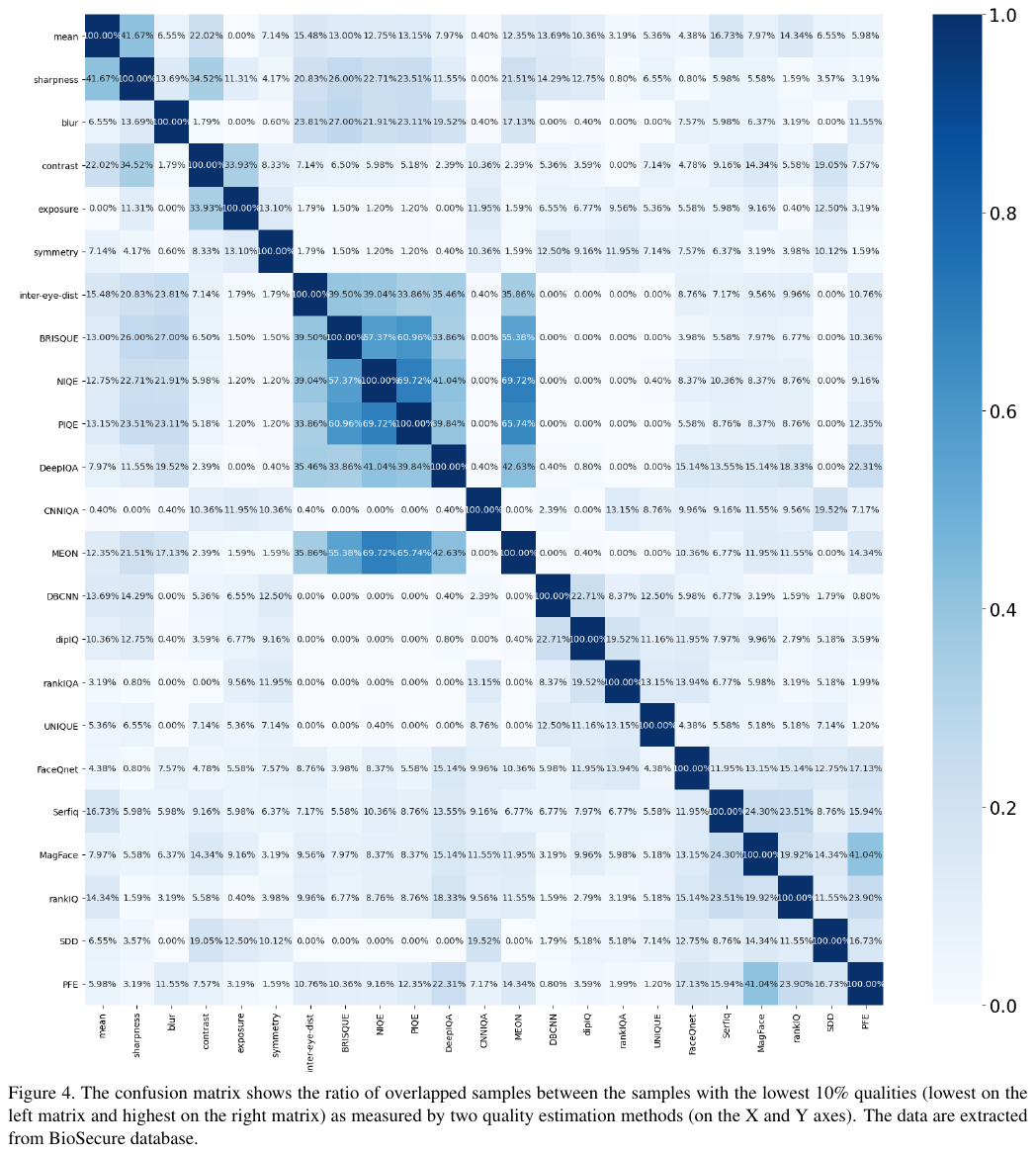}
\end{figure*}

\begin{figure*}
    \centering
    \includegraphics[page=12, width=\linewidth]{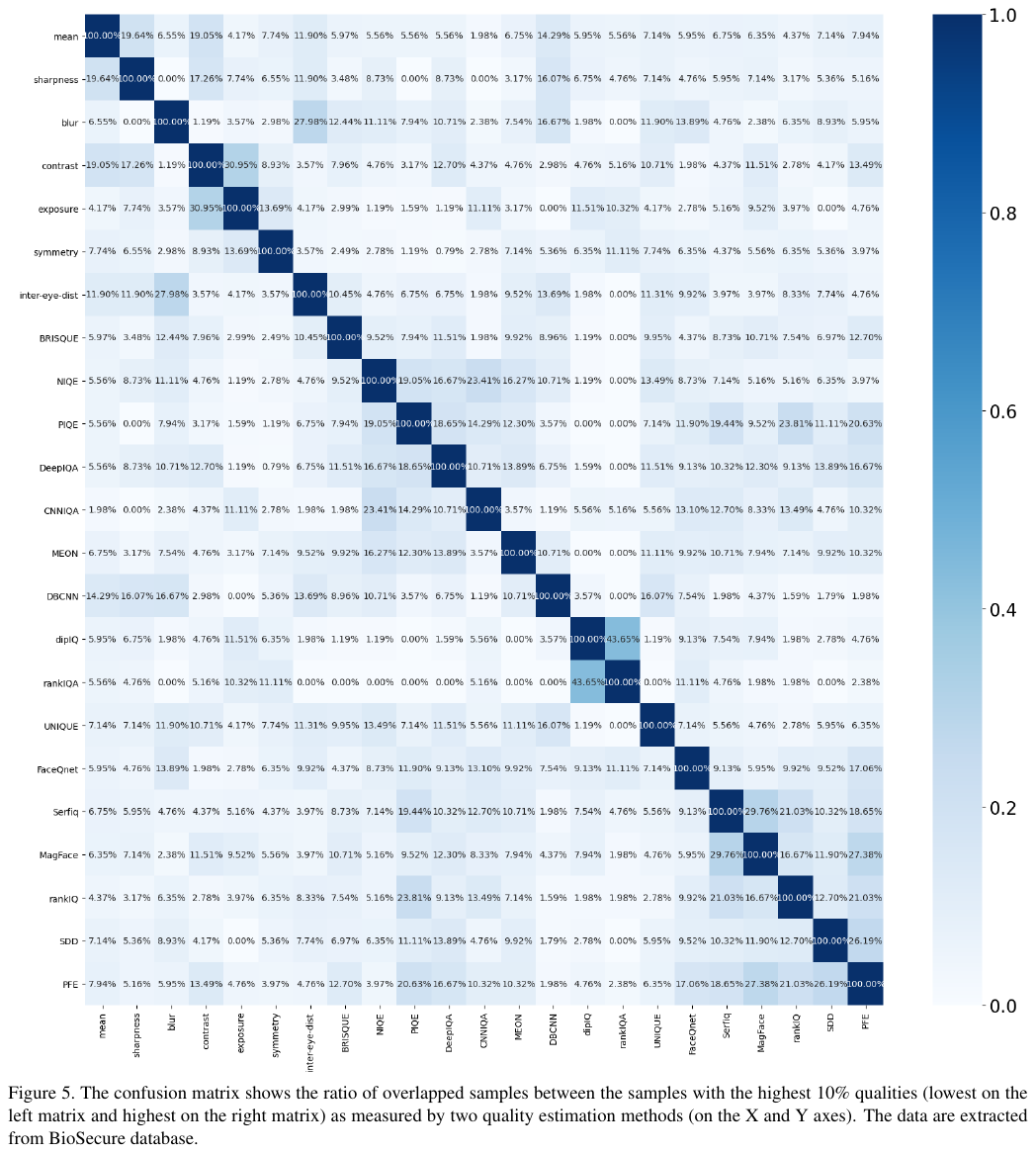}
\end{figure*}

\begin{figure*}
    \centering
    \includegraphics[page=13, width=\linewidth]{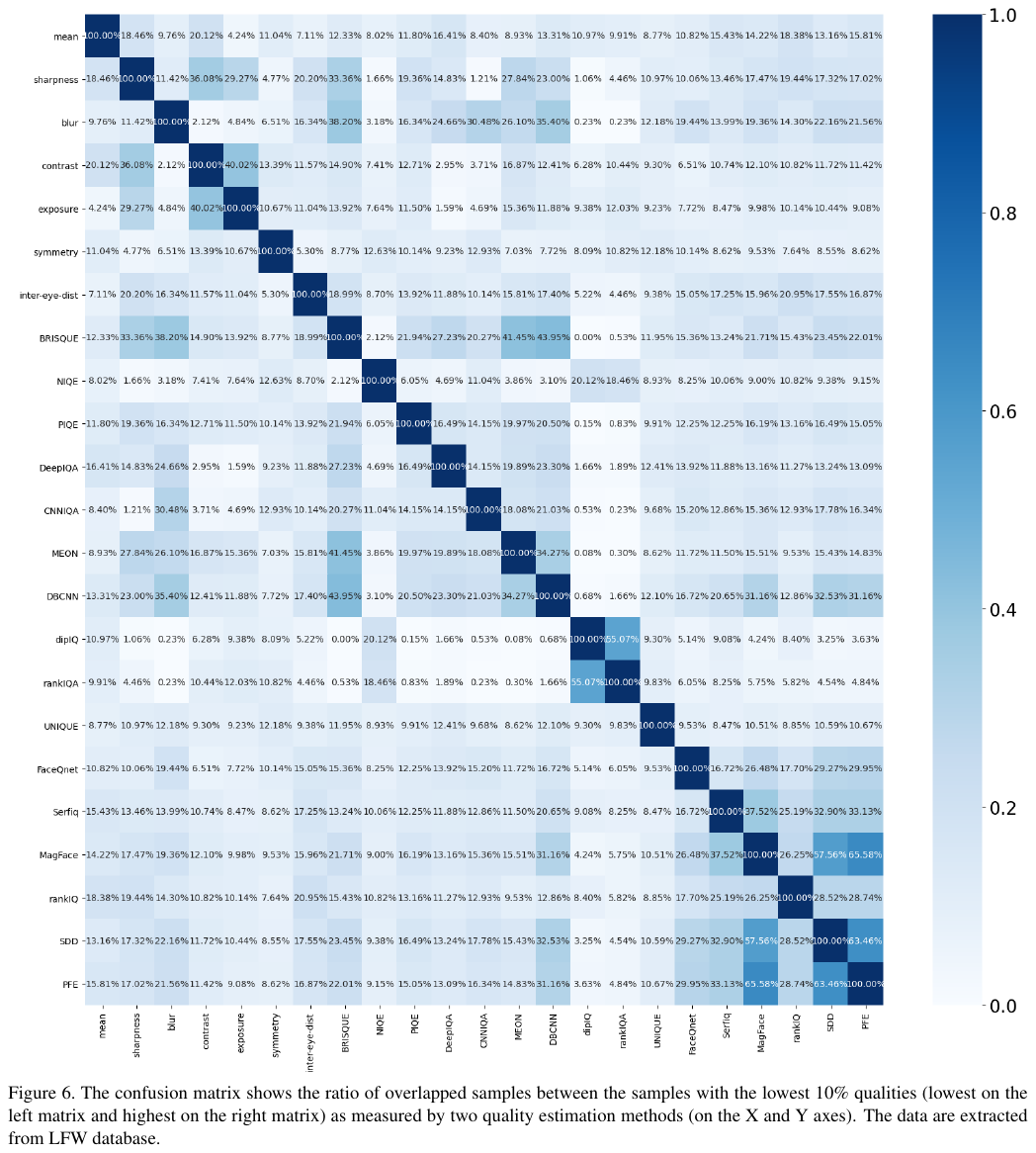}
\end{figure*}

\begin{figure*}
    \centering
    \includegraphics[page=14, width=\linewidth]{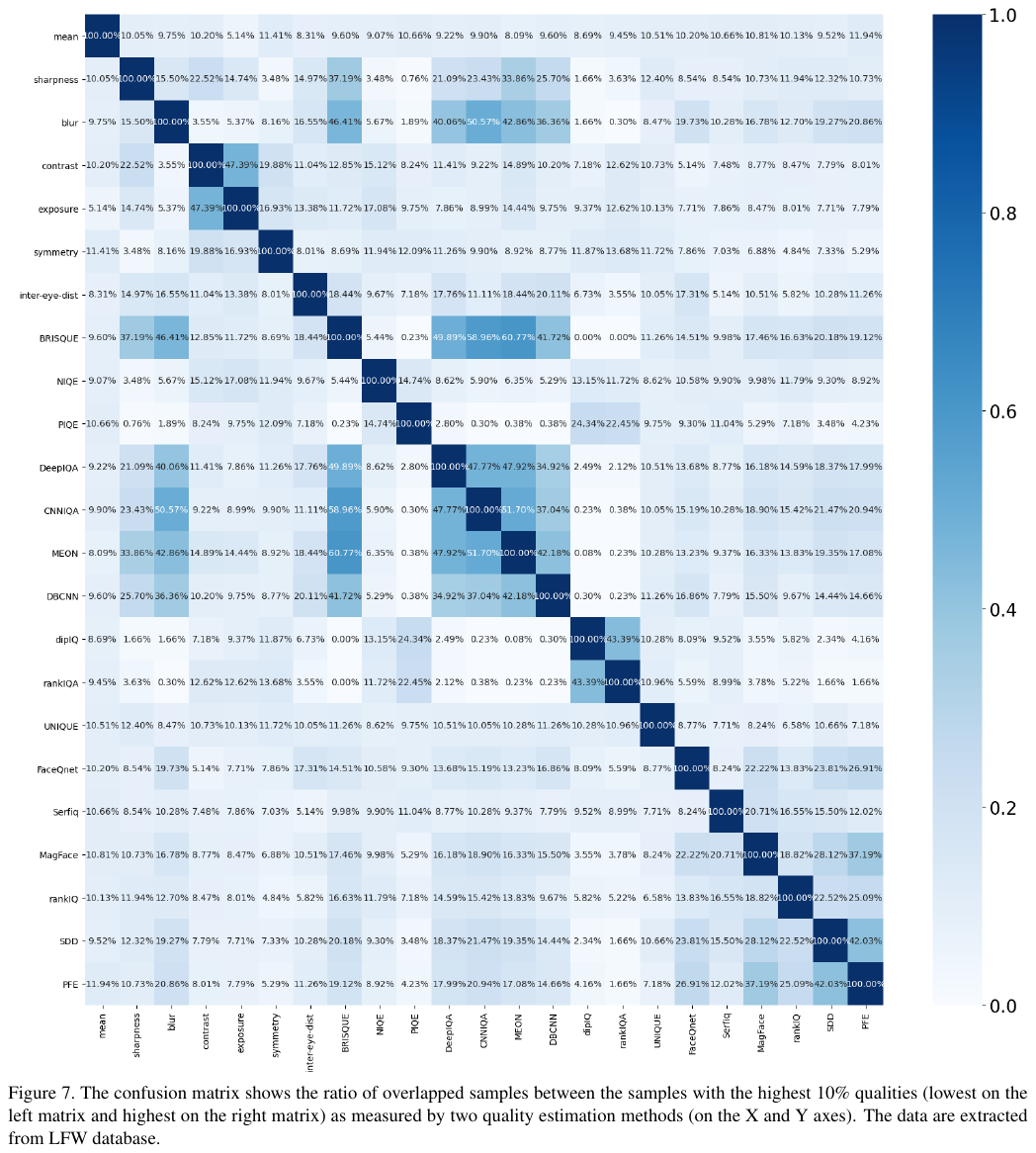}
\end{figure*}

{\small
\bibliographystyle{ieee_fullname}
\bibliography{egbib}

\begin{thebibliography}{10}\itemsep=-1pt

\bibitem{best2018learning}
Lacey Best-Rowden and Anil~K Jain.
\newblock Learning face image quality from human assessments.
\newblock {\em IEEE Transactions on Information Forensics and Security},
  13(12):3064--3077, 2018.

\bibitem{bosse2017deep}
Sebastian Bosse, Dominique Maniry, Klaus-Robert M{\"u}ller, Thomas Wiegand, and
  Wojciech Samek.
\newblock Deep neural networks for no-reference and full-reference image
  quality assessment.
\newblock {\em IEEE TIP}, 27(1):206--219, 2017.

\bibitem{DBLP:journals/corr/abs-2109-09416}
Fadi Boutros, Naser Damer, Florian Kirchbuchner, and Arjan Kuijper.
\newblock Elasticface: Elastic margin loss for deep face recognition.
\newblock {\em CoRR}, abs/2109.09416, 2021.

\bibitem{Cao18}
Q. Cao, L. Shen, W. Xie, O.~M. Parkhi, and A. Zisserman.
\newblock {VGGFace2}: A dataset for recognising faces across pose and age.
\newblock In {\em Int.~Conf.~on Automatic Face and Gesture Recognition}, 2018.

\bibitem{DBLP:conf/icml/Caruana96}
Rich Caruana.
\newblock Algorithms and applications for multitask learning.
\newblock In {\em {ICML}}, pages 87--95. Morgan Kaufmann, 1996.

\bibitem{chen2015face}
Jiansheng Chen, Yu Deng, Gaocheng Bai, and Guangda Su.
\newblock Face image quality assessment based on learning to rank.
\newblock {\em IEEE SPL}, 22(1):90--94, 2015.

\bibitem{DBLP:conf/icpr/DamerSN14}
Naser Damer, Timotheos Samartzidis, and Alexander Nouak.
\newblock Personalized face reference from video: Key-face selection and
  feature-level fusion.
\newblock In {\em Face and Facial Expression Recognition from Real World Videos
  -- Int.~Workshop, FFER@ICPR 2014, Stockholm, Sweden, August 24, 2014, Revised
  Selected Papers}, volume 8912 of {\em Lecture Notes in Computer Science},
  pages 85--98. Springer, 2014.

\bibitem{DBLP:conf/cvpr/DengDSLL009}
Jia Deng, Wei Dong, Richard Socher, Li{-}Jia Li, Kai Li, and Fei{-}Fei Li.
\newblock Imagenet: {A} large-scale hierarchical image database.
\newblock In {\em {CVPR}}, pages 248--255. {IEEE} Computer Society, 2009.

\bibitem{deng2019arcface}
Jiankang Deng, Jia Guo, Niannan Xue, and Stefanos Zafeiriou.
\newblock Arcface: Additive angular margin loss for deep face recognition.
\newblock In {\em Proceedings of the IEEE/CVF CVPR}, pages 4690--4699, 2019.

\bibitem{dutta2014bayesian}
Abhishek Dutta, Raymond Veldhuis, and Luuk Spreeuwers.
\newblock A bayesian model for predicting face recognition performance using
  image quality.
\newblock In {\em IEEE Int.~Joint Conf.~on Biometrics}, pages 1--8. IEEE, 2014.

\bibitem{dutta2015predicting}
Abhishek Dutta, Raymond Veldhuis, and Luuk Spreeuwers.
\newblock Predicting face recognition performance using image quality.
\newblock {\em arXiv preprint arXiv:1510.07119}, 2015.

\bibitem{Fer12}
Matteo Ferrara, Annalisa Franco, Dario Maio, and Davide Maltoni.
\newblock Face image conformance to {ISO/ICAO} standards in machine readable
  travel documents.
\newblock {\em {IEEE} Trans. Inf. Forensics Secur.}, 7(4):1204--1213, 2012.

\bibitem{frontex2015best}
Frontex.
\newblock Best practice technical guidelines for automated border control (abc)
  systems, 2015.

\bibitem{9548297}
Biying Fu, Cong Chen, Olaf Henniger, and Naser Damer.
\newblock The relative contributions of facial parts qualities to the face
  image utility.
\newblock In {\em 2021 International Conference of the Biometrics Special
  Interest Group (BIOSIG)}, pages 1--5, 2021.

\bibitem{9548302}
Biying Fu, Noémie Spiller, Cong Chen, and Naser Damer.
\newblock The effect of face morphing on face image quality.
\newblock In {\em 2021 International Conference of the Biometrics Special
  Interest Group (BIOSIG)}, pages 1--5, 2021.

\bibitem{GT07}
P. Grother and E. Tabassi.
\newblock Performance of biometric quality measures.
\newblock {\em IEEE Trans.~on Pattern Analysis and Machine Intelligence},
  29(4):531--543, Apr. 2007.

\bibitem{guo2016ms}
Yandong Guo, Lei Zhang, Yuxiao Hu, Xiaodong He, and Jianfeng Gao.
\newblock {MS-Celeb-1M}: A dataset and benchmark for large-scale face
  recognition.
\newblock In {\em European Conf.~on Computer Vision}, pages 87--102. Springer,
  2016.

\bibitem{han2017deep}
Dongyoon Han, Jiwhan Kim, and Junmo Kim.
\newblock Deep pyramidal residual networks.
\newblock In {\em {CVPR}}, pages 6307--6315. {IEEE} Computer Society, 2017.

\bibitem{he2016deep}
Kaiming He, Xiangyu Zhang, Shaoqing Ren, and Jian Sun.
\newblock Deep residual learning for image recognition.
\newblock In {\em {CVPR}}, pages 770--778. {IEEE} Computer Society, 2016.

\bibitem{hernandez2020biometric}
Javier Hernandez{-}Ortega, Javier Galbally, Julian Fi{\'{e}}rrez, and Laurent
  Beslay.
\newblock Biometric quality: Review and application to face recognition with
  faceqnet.
\newblock {\em CoRR}, abs/2006.03298, 2020.

\bibitem{hernandez2019faceqnet}
Javier Hernandez-Ortega, Javier Galbally, Julian Fierrez, Rudolf Haraksim, and
  Laurent Beslay.
\newblock {FaceQnet}: Quality assessment for face recognition based on deep
  learning.
\newblock In {\em Int.~Conf.~on Biometrics (ICB)}, pages 1--8. IEEE, 2019.

\bibitem{huang2014labeled}
Gary~B Huang and Erik Learned-Miller.
\newblock Labeled faces in the wild: Updates and new reporting procedures.
\newblock {\em Dept. Comput. Sci., Univ. Massachusetts Amherst, Amherst, MA,
  USA, Tech. Rep}, 14(003), 2014.

\bibitem{LFWTech}
Gary~B. Huang, Manu Ramesh, Tamara Berg, and Erik Learned-Miller.
\newblock Labeled faces in the wild: A database for studying face recognition
  in unconstrained environments.
\newblock Technical Report 07-49, University of Massachusetts, Amherst, Oct.
  2007.

\bibitem{huynh2008scope}
Quan Huynh-Thu and Mohammed Ghanbari.
\newblock Scope of validity of {PSNR} in image/video quality assessment.
\newblock {\em Electronics Letters}, 44(13):800--801, 2008.

\bibitem{ICAO18}
Portrait quality (reference facial images for {MRTD}).
\newblock ICAO Technical Report, 2018.

\bibitem{ISOIEC29794-1}
{Information technology -- Biometric sample quality -- Part~1: Framework}.
\newblock Int.~Standard ISO/IEC 29794-1, 2016.

\bibitem{ISOIEC29794-5}
{Information technology -- Biometric sample quality -- Part~5: Face image
  data}.
\newblock Technical Report ISO/IEC TR 29794-5, 2010.

\bibitem{janssen2018google}
J Janssen.
\newblock {Google Pay startet in Deutschland: Bezahlen mit dem Android-Handy}.
\newblock https://www.heise.de, 2018.

\bibitem{kang2014convolutional}
Le Kang, Peng Ye, Yi Li, and David Doermann.
\newblock Convolutional neural networks for no-reference image quality
  assessment.
\newblock In {\em Proc.~of the IEEE Conf.~on CVPR}, pages 1733--1740, 2014.

\bibitem{li2009reduced}
Qiang Li and Zhou Wang.
\newblock Reduced-reference image quality assessment using divisive
  normalization-based image representation.
\newblock {\em IEEE journal of selected topics in signal processing},
  3(2):202--211, 2009.

\bibitem{liu2019apple}
Stephanie~Q Liu and Anna~S Mattila.
\newblock {Apple Pay}: Coolness and embarrassment in the service encounter.
\newblock {\em Int.~Journal of Hospitality Management}, 78:268--275, 2019.

\bibitem{liu2017sphereface}
Weiyang Liu, Yandong Wen, Zhiding Yu, Ming Li, Bhiksha Raj, and Le Song.
\newblock {SphereFace}: Deep hypersphere embedding for face recognition.
\newblock In {\em Proc.~of the IEEE Conf.~on Computer Vision and Pattern
  Recognition}, pages 212--220, 2017.

\bibitem{liu2017rankiqa}
Xialei Liu, Joost Van De~Weijer, and Andrew~D Bagdanov.
\newblock {RankIQA}: Learning from rankings for no-reference image quality
  assessment.
\newblock In {\em Proc.~of the IEEE Int.~Conf.~on Computer Vision}, pages
  1040--1049, 2017.

\bibitem{7934456}
K. {Ma}, W. {Liu}, T. {Liu}, Z. {Wang}, and D. {Tao}.
\newblock {dipIQ}: Blind image quality assessment by learning-to-rank
  discriminable image pairs.
\newblock {\em IEEE Trans.~on Image Processing}, 26(8):3951--3964, 2017.

\bibitem{ma2017end}
Kede Ma, Wentao Liu, Kai Zhang, Zhengfang Duanmu, Zhou Wang, and Wangmeng Zuo.
\newblock End-to-end blind image quality assessment using deep neural networks.
\newblock {\em IEEE TIP}, 27(3):1202--1213, 2017.

\bibitem{meng2021magface}
Qiang Meng, Shichao Zhao, Zhida Huang, and Feng Zhou.
\newblock Magface: A universal representation for face recognition and quality
  assessment.
\newblock In {\em IEEE Conf.~on Computer Vision and Pattern Recognition}, 2021.

\bibitem{mittal2012no}
Anish Mittal, Anush~Krishna Moorthy, and Alan~Conrad Bovik.
\newblock No-reference image quality assessment in the spatial domain.
\newblock {\em IEEE TIP}, 21(12):4695--4708, 2012.

\bibitem{mittal2012making}
Anish Mittal, Rajiv Soundararajan, and Alan~C Bovik.
\newblock Making a ``completely blind'' image quality analyzer.
\newblock {\em IEEE SPL}, 20(3):209--212, 2012.

\bibitem{ortega2009multiscenario}
Javier Ortega{-}Garcia, Julian Fi{\'{e}}rrez, Fernando Alonso{-}Fernandez,
  Javier Galbally, Manuel~R. Freire, Joaquin Gonzalez{-}Rodriguez, Carmen
  Garc{\'{\i}}a{-}Mateo, Jos{\'{e}}~Luis Alba{-}Castro, Elisardo
  Gonz{\'{a}}lez{-}Agulla, Enrique~Otero Muras, Sonia Garcia{-}Salicetti,
  Lor{\`{e}}ne Allano, Van{-}Bao Ly, Bernadette Dorizzi, Josef Kittler,
  Thirimachos Bourlai, Norman Poh, Farzin Deravi, Ming W.~R. Ng, Michael~C.
  Fairhurst, Jean Hennebert, Andreas Humm, Massimo Tistarelli, Linda Brodo,
  Jonas Richiardi, Andrzej Drygajlo, Harald Ganster, Federico Sukno,
  Sri{-}Kaushik Pavani, Alejandro~F. Frangi, Lale Akarun, and Arman Savran.
\newblock The multiscenario multienvironment biosecure multimodal database
  {(BMDB)}.
\newblock {\em {IEEE} Trans. Pattern Anal. Mach. Intell.}, 32(6):1097--1111,
  2010.

\bibitem{DBLP:journals/corr/abs-2103-05977}
Fu{-}Zhao Ou, Xingyu Chen, Ruixin Zhang, Yuge Huang, Shaoxin Li, Jilin Li, Yong
  Li, Liujuan Cao, and Yuan{-}Gen Wang.
\newblock {SDD-FIQA:} unsupervised face image quality assessment with
  similarity distribution distance.
\newblock {\em CoRR}, abs/2103.05977, 2021.

\bibitem{ponomarenko2015image}
Nikolay Ponomarenko, Lina Jin, Oleg Ieremeiev, Vladimir Lukin, Karen
  Egiazarian, Jaakko Astola, Benoit Vozel, Kacem Chehdi, Marco Carli, Federica
  Battisti, et~al.
\newblock Image database {TID2013}: Peculiarities, results and perspectives.
\newblock {\em Signal Processing: Image Communication}, 30:57--77, 2015.

\bibitem{ponomarenko2009tid2008}
Nikolay Ponomarenko, Vladimir Lukin, Alexander Zelensky, Karen Egiazarian,
  Marco Carli, and Federica Battisti.
\newblock {TID2008} -- a database for evaluation of full-reference visual
  quality assessment metrics.
\newblock {\em Advances of Modern Radioelectronics}, 10(4):30--45, 2009.

\bibitem{rehman2012reduced}
Abdul Rehman and Zhou Wang.
\newblock Reduced-reference image quality assessment by structural similarity
  estimation.
\newblock {\em IEEE transactions on image processing}, 21(8):3378--3389, 2012.

\bibitem{ruderman1994statistics}
Daniel~L Ruderman.
\newblock The statistics of natural images.
\newblock {\em Network: Computation in Neural Systems}, 5(4):517--548, 1994.

\bibitem{schroff2015facenet}
Florian Schroff, Dmitry Kalenichenko, and James Philbin.
\newblock {FaceNet}: A unified embedding for face recognition and clustering.
\newblock In {\em Proc.~of the IEEE Conf.~on Computer Vision and Pattern
  Recognition}, pages 815--823, 2015.

\bibitem{sheikh2005live}
HR Sheikh.
\newblock {LIVE} image quality assessment database release 2.
\newblock \url{http://live.ece.utexas.edu/research/quality}, 2005.

\bibitem{DBLP:conf/iccv/ShiJ19}
Yichun Shi and Anil~K. Jain.
\newblock Probabilistic face embeddings.
\newblock In {\em {ICCV}}, pages 6901--6910. {IEEE}, 2019.

\bibitem{terhorst2020face}
Philipp Terh{\"o}rst, Jan~Niklas Kolf, Naser Damer, Florian Kirchbuchner, and
  Arjan Kuijper.
\newblock Face quality estimation and its correlation to demographic and
  non-demographic bias in face recognition.
\newblock In {\em 2020 IEEE Int.~Joint Conf.~on Biometrics (IJCB)}, pages
  1--11. IEEE, 2020.

\bibitem{terhorst2020ser}
Philipp Terh\"{o}rst, Jan~Niklas Kolf, Naser Damer, Florian Kirchbuchner, and
  Arjan Kuijper.
\newblock {SER-FIQ}: Unsupervised estimation of face image quality based on
  stochastic embedding robustness.
\newblock In {\em Proc.~of the IEEE/CVF Conf.~on Computer Vision and Pattern
  Recognition}, pages 5651--5660, 2020.

\bibitem{venkatanath2015blind}
N Venkatanath, D Praneeth, Maruthi~Chandrasekhar Bh, Sumohana~S Channappayya,
  and Swarup~S Medasani.
\newblock Blind image quality evaluation using perception based features.
\newblock In {\em 21\textsuperscript{st} National Conf.~on Communications
  (NCC)}, pages 1--6. IEEE, 2015.

\bibitem{wang2020score}
Haofan Wang, Zifan Wang, Mengnan Du, Fan Yang, Zijian Zhang, Sirui Ding, Piotr
  Mardziel, and Xia Hu.
\newblock {Score-CAM}: Score-weighted visual explanations for convolutional
  neural networks.
\newblock In {\em Proc.~of the IEEE/CVF Conf.~on Computer Vision and Pattern
  Recognition Workshops}, pages 24--25, 2020.

\bibitem{ssim}
Z Wang, E~P Simoncelli, and A~C Bovik.
\newblock Multiscale structural similarity for image quality assessment.
\newblock In {\em Proc.~37\textsuperscript{th} Asilomar Conf.~on Signals,
  Systems and Computers}, volume~2, pages 1398--1402, Pacific Grove, CA, Nov.
  2003. IEEE Computer Society.

\bibitem{wolf2011face}
Lior Wolf, Tal Hassner, and Itay Maoz.
\newblock Face recognition in unconstrained videos with matched background
  similarity.
\newblock In {\em CVPR 2011}, pages 529--534. IEEE, 2011.

\bibitem{wu2013reduced}
Jinjian Wu, Weisi Lin, Guangming Shi, and Anmin Liu.
\newblock Reduced-reference image quality assessment with visual information
  fidelity.
\newblock {\em IEEE Transactions on Multimedia}, 15(7):1700--1705, 2013.

\bibitem{DBLP:journals/corr/YiLLL14a}
Dong Yi, Zhen Lei, Shengcai Liao, and Stan~Z. Li.
\newblock Learning face representation from scratch.
\newblock {\em CoRR}, abs/1411.7923, 2014.

\bibitem{zhang2016joint}
Kaipeng Zhang, Zhanpeng Zhang, Zhifeng Li, and Yu Qiao.
\newblock Joint face detection and alignment using multitask cascaded
  convolutional networks.
\newblock {\em IEEE Signal Processing Letters}, 23(10):1499--1503, 2016.

\bibitem{DBLP:journals/tcsv/ZhangMYDW20}
Weixia Zhang, Kede Ma, Jia Yan, Dexiang Deng, and Zhou Wang.
\newblock Blind image quality assessment using a deep bilinear convolutional
  neural network.
\newblock {\em {IEEE} Trans. Circuits Syst. Video Technol.}, 30(1):36--47,
  2020.

\bibitem{DBLP:journals/tip/ZhangMZY21}
Weixia Zhang, Kede Ma, Guangtao Zhai, and Xiaokang Yang.
\newblock Uncertainty-aware blind image quality assessment in the laboratory
  and wild.
\newblock {\em {IEEE} Trans. Image Process.}, 30:3474--3486, 2021.

\end{thebibliography}
}

\end{document}